\pdfoutput=1
\documentclass[11pt]{article}

\usepackage[]{acl}

\usepackage{times}
\usepackage{latexsym}
\usepackage{hyperref}
\usepackage{color, colortbl}
\definecolor{Gray}{gray}{0.9}
\usepackage{array}
\usepackage{booktabs}
\usepackage{multirow}
\usepackage{amsmath}
\usepackage{amssymb}
\usepackage{graphicx}
\usepackage{enumerate}
\usepackage{diagbox}
\usepackage{mathtools}
\usepackage{paralist}
\usepackage{tabularx}
\usepackage{lipsum}
\usepackage{ragged2e}
\usepackage{makecell}

\usepackage{xcolor}
\usepackage{color, colortbl}

\definecolor{colorxmark}{RGB}{255, 87, 51}
\definecolor{colorcmark}{RGB}{66, 154, 137}
\definecolor{headcolor}{HTML}{018161}
\definecolor{relationcolor}{HTML}{d95f02}
\definecolor{tailcolor}{HTML}{6560a3}
\definecolor{concept_color}{HTML}{385723}
\definecolor{instance_color}{HTML}{1F4E79}
\definecolor{new_knowledge_color}{HTML}{7030A0}
\definecolor{original_tail_color}{HTML}{BF9000}

\usepackage[T1]{fontenc}

\usepackage[utf8]{inputenc}
\usepackage{csquotes}

\usepackage{microtype}

\usepackage{inconsolata}

\newcommand{\candle}{\raisebox{-3.5pt}{\includegraphics[width=1.2em]{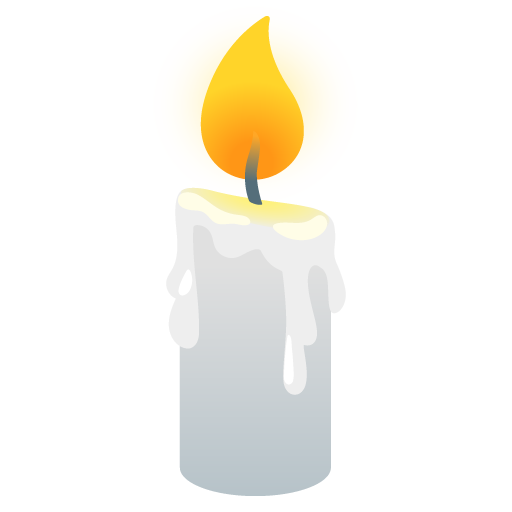}}}

\newcommand*{\affaddr}[1]{#1} 
\newcommand*{\affmark}[1][*]
{\textsuperscript{#1}}

\title{\candle\text{CANDLE}: Iterative Conceptualization and Instantiation Distillation from Large Language Models for Commonsense Reasoning}

\author{
Weiqi Wang\affmark[1],
Tianqing Fang\affmark[1],
Chunyang Li\affmark[2],
Haochen Shi\affmark[1],
Wenxuan Ding\affmark[1],
Baixuan Xu\affmark[1],\\
\textbf{
Zhaowei Wang\affmark[1],
Jiaxin Bai\affmark[1],
Xin Liu\affmark[3],
Jiayang Cheng\affmark[1],
Chunkit Chan\affmark[1],
Yangqiu Song\affmark[1]}\\
\affaddr{\affmark[1]Department of Computer Science and Engineering, HKUST, Hong Kong SAR, China}\\
\affaddr{\affmark[2]Department of Computer Science and Technology, Tsinghua Univerisity, Beijing, China}\\
\affaddr{\affmark[3]Amazon.com Inc, Palo Alto, USA}\\
\texttt{\{wwangbw, tfangaa, yqsong\}@cse.ust.hk}\\ 
}

\begin{document}
\maketitle
\begin{abstract}
The sequential process of conceptualization and instantiation is essential to generalizable commonsense reasoning as it allows the application of existing knowledge to unfamiliar scenarios. 
However, existing works tend to undervalue the step of instantiation and heavily rely on pre-built concept taxonomies and human annotations to collect both types of knowledge, resulting in a lack of instantiated knowledge to complete reasoning, high cost, and limited scalability.
To tackle these challenges, we introduce CANDLE (\underline{\textbf{C}}onceptu\underline{\textbf{A}}lization and I\underline{\textbf{N}}stantiation \underline{\textbf{D}}istillation from \underline{\textbf{L}}arge Language Mod\underline{\textbf{E}}ls), a distillation framework that iteratively performs contextualized conceptualization and instantiation over commonsense knowledge bases by instructing large language models to generate both types of knowledge with critic filtering. 
By applying CANDLE to ATOMIC~\cite{DBLP:conf/aaai/SapBABLRRSC19}, we construct a comprehensive knowledge base comprising six million conceptualizations and instantiated commonsense knowledge triples. 
Both types of knowledge are firmly rooted in the original ATOMIC dataset, and intrinsic evaluations demonstrate their exceptional quality and diversity.
Empirical results indicate that distilling CANDLE on student models provides benefits across three downstream tasks\footnote{Our data and models are publicly available at \href{https://github.com/HKUST-KnowComp/CANDLE}{https://github.com/HKUST-KnowComp/CANDLE}.}.
\end{abstract}

\section{Introduction}

\begin{figure}[t]
     \centering
     \includegraphics[width=1\linewidth]{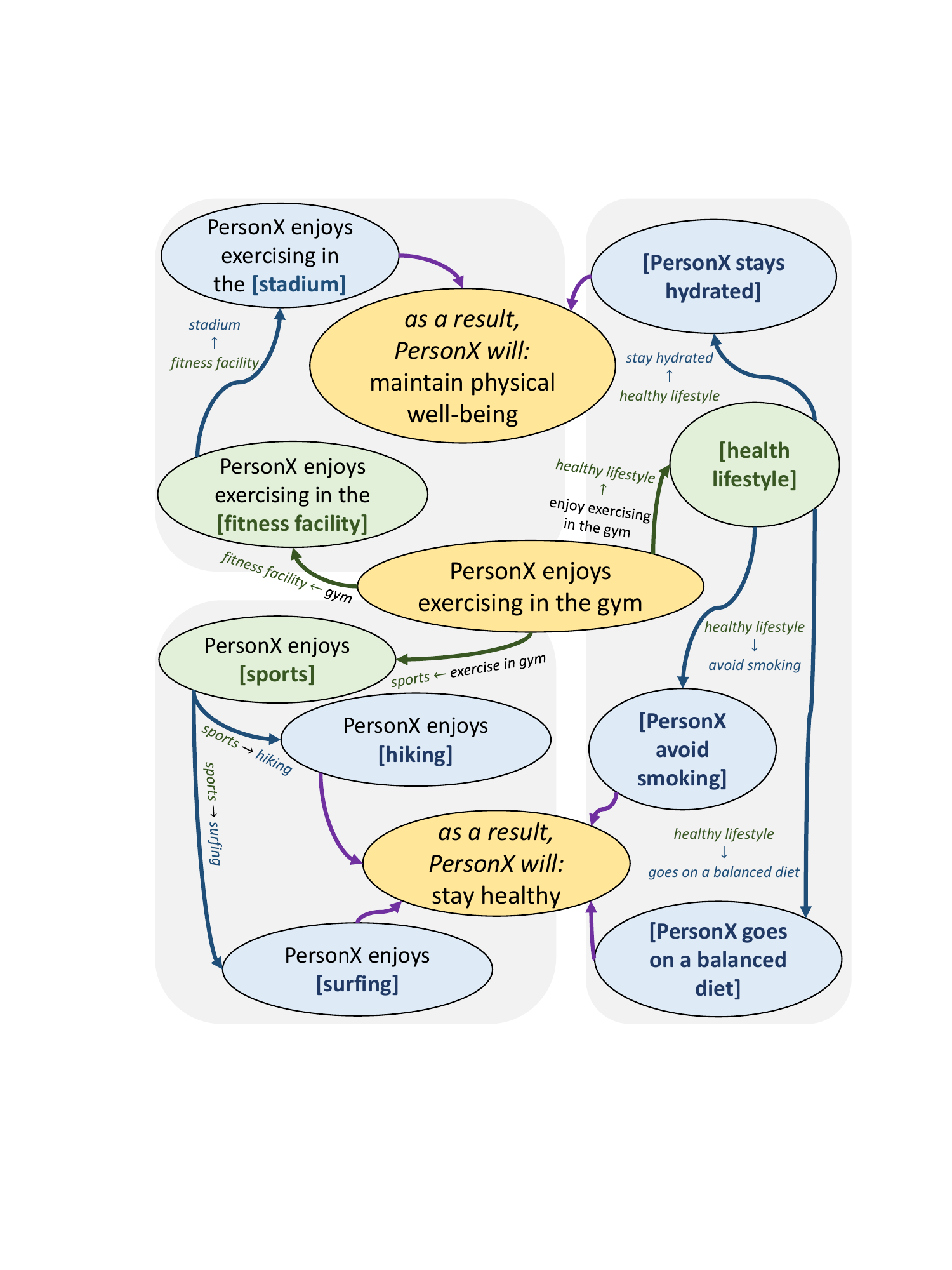}
     \vspace{-0.1in}
     \caption{Examples showing several chains of \textbf{\textcolor{concept_color}{conceptualization}} and \textbf{\textcolor{instance_color}{instantiation}} over the event \textit{PersonX enjoys exercising in the gym}.
     \textbf{\textcolor{new_knowledge_color}{New inferential commonsense knowledge}} can be induced when placing the \textbf{\textcolor{instance_color}{instantiation}} back into the \textbf{\textcolor{original_tail_color}{original context}}.}
    \label{fig:introduction}
    \vspace{-0.1in}
\end{figure}

Commonsense reasoning refers to the cognitive ability to make logical inferences and draw conclusions based on general knowledge and understanding of the world that is typically shared among individuals~\cite{davis2014representations,mueller2014commonsense}.
However, a longstanding challenge is generalizability, as commonsense reasoning often necessitates applying knowledge to novel situations beyond simple pattern recognition or memorizing all special cases~\cite{mortimer1995conceptual,banaji1989bankruptcy}.
One promising approach to address this is the chain of conceptualization~\cite{murphy2004big} and instantiation~\cite{anderson1976instantiation}, which, akin to the process of conceptual induction and deduction in human reasoning~\cite{tenenbaum2011grow}, involves conceptualizing instances derived from known commonsense knowledge and subsequently instantiating these concepts in new situations to obtain the knowledge required for downstream reasoning.
For example, in Figure~\ref{fig:introduction}, one can first conceptualize \textit{enjoys exercising in the gym} as a \textit{\textcolor{concept_color}{healthy lifestyle}}, and then further instantiate it to \textit{\textcolor{instance_color}{go on a balanced diet}}. 
This process allows for the derivation of a novel event, \textit{\textcolor{instance_color}{PersonX goes on a balanced diet}}, which may entail \textcolor{new_knowledge_color}{new commonsense knowledge} when connected with the \textcolor{original_tail_color}{original event's commonsense inferential tail}.
By possessing substantial knowledge to initiate the process of conceptualization and instantiation, one can extrapolate limited commonsense knowledge to a wide array of diverse scenarios.

Yet, replicating this fundamental ability on machines remains challenging due to the absence of both types of knowledge in widely used CommonSense Knowledge Bases (CSKBs;~\citealp{DBLP:conf/aaai/SapBABLRRSC19,DBLP:conf/aaai/SpeerCH17,DBLP:conf/www/FangZWSH21,DBLP:conf/emnlp/FangWCHZSH21}).
Various methods compensating the lack of conceptualization ability of language models have been proposed for entity-level~\cite{DBLP:conf/eacl/DurmeMS09,DBLP:conf/ijcai/SongWWLC11,DBLP:conf/ijcai/SongWW15,DBLP:conf/aaai/GongZZ16,DBLP:journals/corr/abs-2003-03239,DBLP:conf/emnlp/PengWHJ0L0022} and event-level~\cite{DBLP:conf/conll/ChenZWR20,AbstractATOMIC,CAT} conceptualizations by matching against concept taxonomies like Probase~\cite{DBLP:conf/sigmod/WuLWZ12} and WordNet~\cite{DBLP:journals/cacm/Miller95}.
However, several limitations still persist.

Firstly, despite the importance of both conceptualization and instantiation, most existing works underestimate the importance of the second step while focusing solely on conceptualization and using the resulting abstract knowledge directly. 
Other studies that concentrate on instantiations either overlook the conceptualization step entirely or only retrieve instances from the original CSKB, failing to introduce novel entities and events.
Secondly, most conceptualization methods heavily depend on matching instances with concepts in concept taxonomies, such as Probase and WordNet, which have a limited scope and lack contextual information. 
Consequently, the derived conceptualizations are constrained in scale by these taxonomies and are formulated without considering proper contextualization, necessitating further verification in the original context.
Lastly, the chain of conceptualization and instantiation can easily bring more than two orders of magnitude of data on top of the original CSKB. 
However, current acquisition and verification methods for both steps heavily rely on human annotation, which can be extremely costly as the scale of the CSKB increases.

To address these gaps, we introduce CANDLE, a \underline{\textbf{C}}onceptu\underline{\textbf{A}}lization and I\underline{\textbf{N}}stantiation \underline{\textbf{D}}istillation framework from \underline{\textbf{L}}arge Language Mod\underline{\textbf{E}}ls (LLMs) to aid commonsense reasoning. 
Specifically, CANDLE marks the first to complete the chain of conceptualization and instantiation by instructing powerful LLMs to sequentially generate both types of knowledge based on concrete commonsense triples while carefully considering the original context throughout the process. 
We further alleviate the human annotation cost by employing two critic filtering models to eliminate low-quality generations.
The instantiated knowledge, representing concrete commonsense knowledge again, can be fed back into CANDLE as input, iteratively augmenting the original CSKB significantly.

By applying CANDLE to ATOMIC~\cite{DBLP:conf/aaai/SapBABLRRSC19}, we construct a large-scale knowledge base comprising 6.18 million conceptualizations and instantiations from two powerful LLMs, ChatGPT~\cite{openai2022chatgpt} and LLAMA2~\cite{LLAMA2}. 
We demonstrate the intrinsic efficacy of CANDLE through automatic and human evaluations, highlighting the ability to generate high-quality and diverse knowledge (Section~\ref{sec:distillation_evaluation}). 
We further show the extrinsic benefits of CANDLE by leveraging the generated knowledge as complementary training data to distill student models that yield improvements across three downstream tasks, including CSKB conceptualization, generative commonsense inference, and zero-shot commonsense question answering (Section~\ref{sec:downstream_applications}). 

\begin{figure*}[t]
     \centering
     \includegraphics[width=1\linewidth]{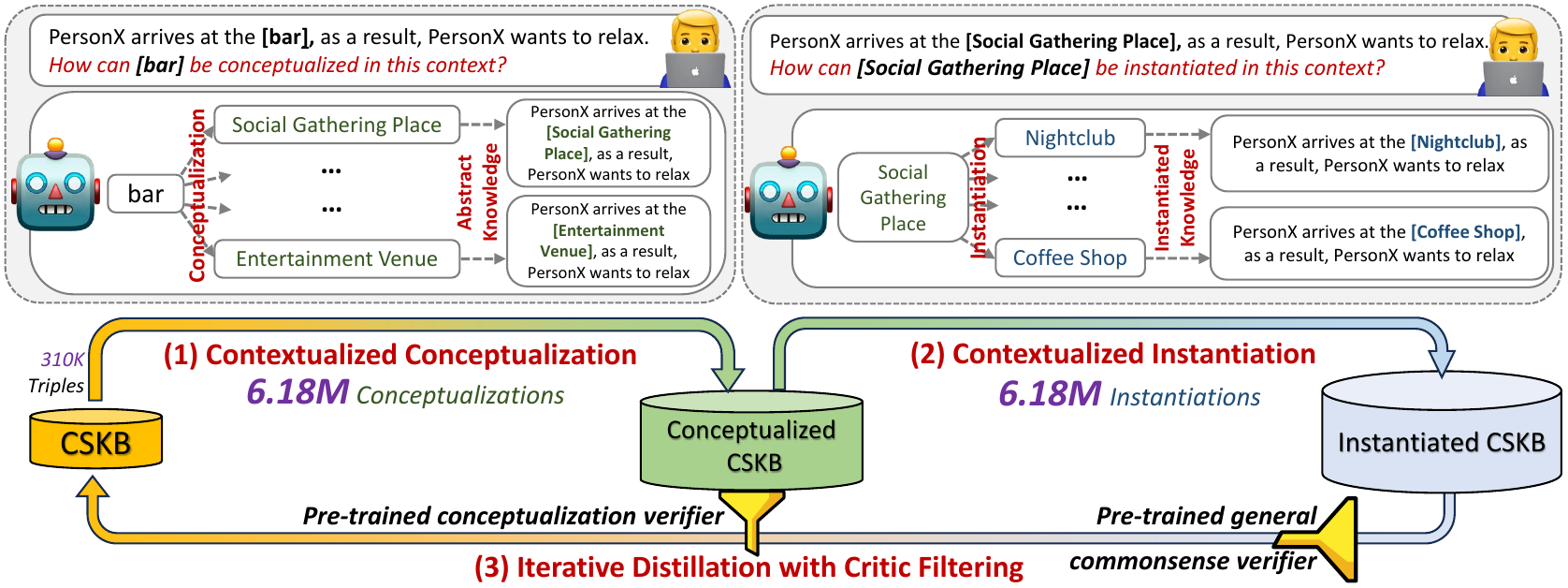}
     \vspace{-0.1in}
     \caption{Overview of our CANDLE framework. 
     A running example with \textit{PersonX arrives at the bar, as a result, PersonX wants to relax} is shown in the figure, where \textit{bar} is first conceptualized and then instantiated by LLMs.
     The instantiations can be integrated back into the original CSKB and become input for the framework again.}
    \label{fig:CANDLE_overview}
\end{figure*}

\section{Related Works}
\subsection{Conceptualization and Instantiation}
Conceptualization aims to abstract a set of entities or events into a general concept, thereby forming abstract commonsense knowledge within its original context~\cite{murphy2004big}. 
Subsequently, instantiation grounds the derived concept into other instances and events to introduce new commonsense knowledge. 
Existing works primarily focused on entity-level conceptualization~\cite{DBLP:conf/eacl/DurmeMS09,DBLP:conf/ijcai/SongWWLC11,DBLP:conf/ijcai/SongWW15,DBLP:journals/ai/LiuCWLCXCJ22,DBLP:conf/emnlp/PengWHJ0L0022}, with~\citet{AbstractATOMIC} pioneering the construction of an event conceptualization benchmark by extracting concepts for social events from WordNet~\cite{DBLP:journals/cacm/Miller95} synsets and Probase~\cite{DBLP:conf/sigmod/WuLWZ12}.  
\citet{CAT,CAR} further proposed a semi-supervised framework for conceptualizing CSKBs and demonstrated that abstract knowledge can enhance commonsense inference modeling and question answering. 
\citet{AbsPyramid} constructed an abstraction benchmark based on eventualities from ASER~\cite{DBLP:journals/ai/ZhangLPKOFS22}. 
Regarding instantiation,~\citet{DBLP:conf/eacl/AllawayHBMDC23} introduced a controllable generative framework to identify valid instantiations for abstract knowledge automatically. 
However, none of the existing studies have fully completed the chain of conceptualization and instantiation, with each focusing on only one aspect. 
Human annotation is also frequently applied for data collection and verification, which is both expensive and limited in scalability.
Additionally, the downstream benefits of instantiated commonsense knowledge have not been thoroughly explored, leaving a significant gap in improving commonsense reasoning models.

\subsection{Commonsense Knowledge Distillation}
Recent breakthroughs in LLMs~\cite{openai2022chatgpt,GPT4} have led to numerous efforts in distilling commonsense knowledge into datasets for training performant student models.
\citet{DBLP:conf/naacl/WestBHHJBLWC22,DBLP:conf/emnlp/SclarWKTC22,DBLP:conf/acl/BhagavatulaHD0L23,DBLP:conf/emnlp/West0SLJLCHBB023} followed the pipeline of symbolic knowledge distillation, which uses human-crafted prompts to extract specific types of knowledge from LLMs for training downstream models.
\citet{DBLP:conf/emnlp/0003GJQ00ZCYD22} proposed to transfer distilled knowledge from a ranker to a retriever, resulting in a more robust commonsense generator.
\citet{DBLP:conf/emnlp/ChaeSOKKYLKY23} and~\citet{DBLP:conf/emnlp/0002HJWLYZ0AKS023} focused on distilling conversational responses from LLMs to enhance dialogue agents with commonsense knowledge and high-quality rationales.
In this paper, we share similar aspirations and propose a chain of distillation framework that sequentially obtains abstract and instantiated knowledge from powerful LLMs. 
Empirical results show that our framework offers more substantial downstream benefits than traditional symbolic knowledge distillation methods.

\section{Definitions and Datasets}
We follow the definitions proposed by~\citet{AbstractATOMIC} and~\citet{CAT} to formulate conceptualization and instantiation. 
Denote the triples in the original CSKB as $D_o=\{(h_o,r,t)|h_o \in H_o, r \in R, t \in T\}$, where $H_o$, $R$, and $T$ are the set of heads, relations, and tails in the original CSKB.
The objective of conceptualization is to form a conceptualized head event, denoted as $h_a$, from the original head $h_o$. 
This is achieved by linking a component $i \subseteq h_o$ to a concept $c$, forming $h_a$ by replacing $i$ with $c$.
Consequently, abstract knowledge is formed by combining the conceptualized head event with the original relation and tail, represented by $(h_a, r, t)$.
In the next step, the goal of instantiation is to associate the concept $c\subseteq h_a$ with a new instance $i'$. 
This process enables the formation of new commonsense knowledge in the format of $(h_{i'}, r, t)$, where $h_{i'}$ is obtained by replacing $c\subseteq h_a$ with $i'$.
In this paper, we use ATOMIC~\cite{DBLP:conf/aaai/SapBABLRRSC19} as the original CSKB $D_o$, which contains 310K $(h_o,r,t)$ triples after dropping those with wildcards and 18,839 unique $h_o$ head events.
AbstractATOMIC~\cite{AbstractATOMIC} is used as the source of instances $i$ for every head event $h_o$.

\section{CANDLE}
This section introduces our CANDLE framework, illustrated in Figure~\ref{fig:CANDLE_overview}.
Our framework can be outlined in three steps:
(1) Instruct ChatGPT to generate contextualized conceptualizations based on the triples in the original CSKB.
(2) Instruct LLAMA2 to instantiate the conceptualizations obtained in Step 1.
(3) Apply critic-filtering to the generations in both steps and close the loop by reintroducing the instantiations back to the CSKB.

\subsection{Contextualized Conceptualization}
\label{sec:contextualized_conceptualization}
Previous methods for collecting conceptualizations rely on heuristically matching instances against concepts from WordNet and Probase. 
However, they suffer from limited concept coverage, resulting in a lack of knowledge diversity after instantiation, and require additional verification to ensure that concept $c$ fits into the original context $(h_o,r,t)$. 
To address both issues, we propose to utilize ChatGPT as a loose teacher to collect conceptualizations in a one-step inference manner.
To verify the feasibility of such choice instead of other open-source LLMs, we carry out a pilot study, in which we randomly select 1000 events and asked ChatGPT and LLAMA2-7B to generate their conceptualizations. 
The results of our expert evaluation show that ChatGPT has 98\% plausible generations, while LLAMA2 only achieves 81\%. 
Therefore, we choose ChatGPT as our core conceptualizer due to its exceptional performances.
Following~\citet{DBLP:conf/nips/BrownMRSKDNSSAA20} and~\citet{DBLP:conf/naacl/WestBHHJBLWC22}, we use a few-shot prompt to instruct ChatGPT:
\begin{center}
\resizebox{0.7\linewidth}{!}{
\begin{tabular}{l}
  \textbf{\texttt{ <TASK-PROMPT> }} \\
  \textbf{\texttt{ \textcolor{headcolor}{<EX$_1$-INP>}\textcolor{tailcolor}{<EX$_1$-OUT>} }}
  \\
  ~~\ldots \\
  \textbf{\texttt{ \textcolor{headcolor}{<EX$_{N-1}$-INP>}\textcolor{tailcolor}{<EX$_{N-1}$-OUT>} }}\\
  \textbf{\texttt{ \textcolor{headcolor}{<EX$_N$-INP>} }}
\end{tabular}
}
\end{center}
where \textbf{\texttt{<TASK-PROMPT>}} is a task instruction that explains how to conceptualize an event and \textbf{\texttt{\textcolor{headcolor}{<EX$_1$-INP>}\textcolor{tailcolor}{<EX$_1$-OUT>}}} are human authored examples of conceptualizations for events sampled from ATOMIC.
For each example, $(h_o,r,t,i)$ are included in the input, and $c$ is the output.
Finally, we provide the N$_{\text{th}}$ input as \textbf{\texttt{\textcolor{headcolor}{<EX$_N$-INP>}}} and ask ChatGPT to generate the corresponding conceptualization as \textbf{\texttt{\textcolor{tailcolor}{<EX$_N$-OUT>}}}.
This ensures that ChatGPT not only learns the relationship between instances $i$ and their conceptualizations $c$ but also performs such abstraction in a contextualized manner, ensuring the plausibility of the generated conceptualization $c$ within the original context $(h_o,r,t)$. 
In this paper, we set $N$ = 6 and obtain $N_c$ = 20 conceptualizations for every event $h_o$.

\subsection{Contextualized Instantiation}
\label{sec:contextualized_instantiation}
After conceptualizing all events, we proceed to instantiate them by instructing an open-source LLM to reduce the cost as the scale of instantiation is $N_c$ = 20 times larger than that of conceptualization.
Similarly, we carry out a round of pilot study to demonstrate the feasibility of employing LLAMA2.
We ask both ChatGPT and LLAMA2 to instantiate 1,000 ChatGPT-generated conceptualizations, and find that both models are able to produce approximately 95\% plausible instantiations with critic filtering. 
Considering the significant cost of using ChatGPT to generate 6.18 million conceptualizations, we decide to use LLAMA2-13B as our core instantiater (Case studies are shown in Appendix~\ref{appendix:case_study}). 
We employ a similar prompt as described in Section~\ref{sec:contextualized_conceptualization}, with the modification of replacing \textbf{\texttt{<TASK-PROMPT>}} with the explanation of instantiating a conceptualized event and changing \textbf{\texttt{\textcolor{headcolor}{<EX$_1$-INP>}\textcolor{tailcolor}{<EX$_1$-OUT>}}} to human-authored examples of instantiations for abstract commonsense knowledge triples. 
$(h_a,r,t,c)$ are included in the input and $i'$ is the expected output.
By learning from these examples, LLAMA2 is expected to generate the corresponding instantiation $i'$ (\textbf{\texttt{\textcolor{tailcolor}{<EX$_N$-OUT>}}}) based on the given abstract knowledge triple $(h_a, r, t, c)$ (\textbf{\texttt{\textcolor{headcolor}{<EX$_N$-INP>}}}).
We set $N$ = 11 and produce only one instantiation for each conceptualized event $h_a$ due to the significant amount of conceptualizations obtained in the previous step.
Appendix~\ref{appendix:distillation_prompts} provides more details regarding the distillation process.

\subsection{Iterating with Critic Filtering}
\label{sec:iteration_critic_filtering}
Following~\citet{DBLP:conf/naacl/WestBHHJBLWC22}, we use critic filtering models to eliminate low-quality generations from LLMs.
Specifically, we utilize a DeBERTa-v3-large conceptualization discriminator, provided by~\citet{CAT}, and VERA-T5-xxl, provided by~\citet{VERA}, to evaluate the quality of the generated conceptualizations and instantiations, respectively.
We set an empirical threshold value $t$ to serve as the cutoff point for discarding generations with scores below $t$. 
In Section~\ref{sec:distillation_evaluation}, we present evaluations conducted to determine the optimal value for $t$. 
For all downstream applications, we set $t$ = 0.9.
Post-filtering, the instantiated triples $(h_{i'},r,t)$ can be reintroduced as the input for conceptualizations again as they continue to represent concrete commonsense knowledge.
This iterative process of conceptualization and instantiation forms a loop, which enables continuously augmenting a CSKB.
In this paper, we execute the loop only once, but multiple iterations hold the promise of significantly enhancing the CSKB's knowledge coverage.

\section{Evaluations and Analysis}
In this section, we evaluate CANDLE from both intrinsic and extrinsic perspectives.
Intrinsically, we demonstrate the high quality and diversity of conceptualizations and instantiations generated by CANDLE  (Section~\ref{sec:distillation_evaluation}).
Extrinsically, we explore the benefits by applying the distilled knowledge to downstream tasks (Section~\ref{sec:downstream_applications}).

\begin{table}[t]
\setlength{\tabcolsep}{4pt}
\small
\centering
\resizebox{\linewidth}{!}{
\begin{tabular}{@{}l|cc|cc@{}}
\toprule
\multirow{2}{*}{Corpus} & \multicolumn{2}{c|}{Conceptualization} & \multicolumn{2}{c}{Instantiation} \\ 
\cmidrule(l){2-5} 
 & Size (Unq.)/K & Accept & Size (Unq.)/K & Accept \\ 
\midrule
AbsATM & 503.5 (31.22) & - & None & - \\
EXEM & 0.650 (0.650) & - & 25.12 (25.12) & - \\
\midrule
CANDLE & 6,181 (853.5) & 82.6\% & 6,181 (676.7) & 77.9\% \\
(critic$_{\text{0.5}}$) & 4,002 (498.4) & 88.1\% & 4,176 (512.7) & 84.4\% \\
(critic$_{\text{0.7}}$) & 3,272 (382.2) & 93.5\% & 3,098 (455.9) & 89.1\% \\ 
(critic$_{\text{0.9}}$) & 2,137 (219.4) & 97.2\% & 2,208 (382.1) & 94.5\% \\ 
\bottomrule
\end{tabular}
}
\vspace{-0.1in}
\caption{Statistics and expert acceptance rates of CANDLE in comparison to AbstractATOMIC (AbsATM;~\citealp{AbstractATOMIC}) and Exemplar (EXEM;~\citealp{DBLP:conf/eacl/AllawayHBMDC23}). 
Unq stands for unique.}
\vspace{-0.25in}
\label{tab:candle_statistics}
\end{table}

\subsection{Distillation Evaluations}
\label{sec:distillation_evaluation}
\paragraph{Statistics and Quality.}
We present CANDLE distillation statistics based on ATOMIC in Table~\ref{tab:candle_statistics}, showing its superiority in scale and concept coverage compared to other benchmarks. 
Even with a strict critic filtering threshold ($t$ = 0.9), CANDLE maintains its leading position, having the highest count of total and unique knowledge for both types. 
To assess the quality of the distilled knowledge, we recruit four expert annotators to conduct human evaluations on the plausibility of the generated conceptualizations and instantiations.
They are asked to annotate the plausibility of 3,000 randomly sampled abstract commonsense triples $(h_a,r,t)$ and 3,000 instantiated triples $(h_{i'},r,t)$ from the distilled knowledge set. 
Accepted triples are those deemed plausible by all annotators. 
We then analyze accepted triple ratios for different levels of critic filtering, as shown in Table~\ref{tab:candle_statistics}. 
Our findings show that LLMs have impressive conceptualization and instantiation abilities, with initial plausibility rates of 82.6\% and 77.9\% for both types of knowledge, respectively. 
Critic filtering improves plausibility by up to 14.6\% and 16.6\%, demonstrating the effectiveness of our measures in maintaining high-quality distilled knowledge. 
For more annotation details, refer to Appendix E.

\begin{figure}[t]
     \centering
     \includegraphics[width=1\linewidth]{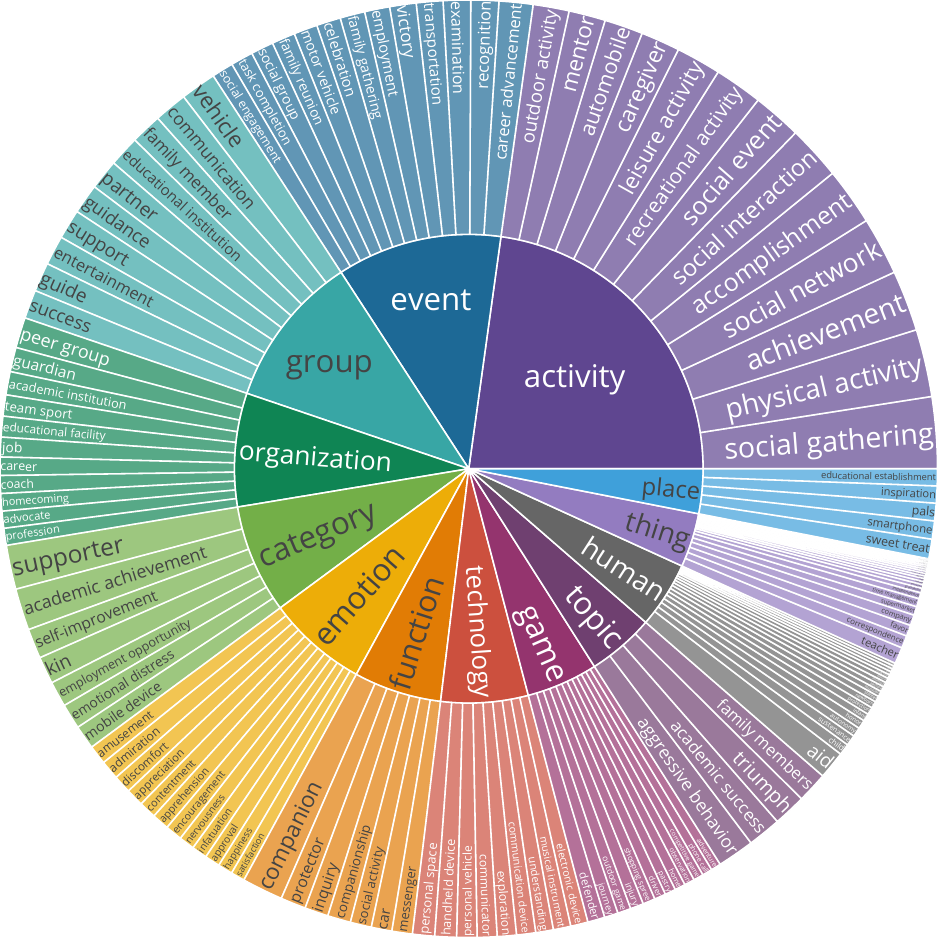}
     \vspace{-0.2in}
     \caption{Hypernyms distribution of the top 10,000 popular conceptualizations distilled from CANDLE.}
    \label{fig:concept_distribution}
    \vspace{-0.2in}
\end{figure}

\vspace{-0.1in}
\paragraph{Conceptualization Diversity.}
The process of abstracting an event into highly diverse conceptualizations plays a crucial role in CANDLE. 
It is of significant importance because the greater the diversity of conceptualizations, the broader the knowledge coverage becomes upon instantiation. 
This, in turn, enhances the overall knowledge coverage within the distillation process.
To examine the diversity of the top 10,000 popular distilled conceptualizations, we obtain their hypernyms by matching against Probase~\cite{DBLP:conf/sigmod/WuLWZ12} and present a visualization in Figure~\ref{fig:concept_distribution}.
It reveals that our distilled conceptualizations possess a high level of diversity across various categories, forming a comprehensive and intricate knowledge base. 
Novelty of our distilled conceptualizations and instantiations, compared to other available resources, are in Appendix~\ref{appendix:distillation_prompts}.

\begin{table*}[t]
\small
\centering
\begin{tabular}{l|l|ll|ll}
\toprule
\multirow{2}{*}{Model Type} & \multirow{2}{*}{Backbone Model / Method} & \multicolumn{2}{c|}{Event Conceptualization} & \multicolumn{2}{c}{Triple Conceptualization} \\
\cmidrule(l){3-6}
 &  & Validation & Testing & Validation & Testing \\ 
\midrule
\multirow{7}{*}{\begin{tabular}[c]{@{}l@{}}Pre-trained\\Langauge\\ Models\end{tabular}} & RoBERTa-large \scriptsize{\textit{340M}} & 77.28 & 77.99 & 81.77 & 82.69 \\
& DeBERTa-v3-large \scriptsize{\textit{435M}} & 78.02 & 78.27 & 82.18 & 82.96 \\
& GPT2-XL \scriptsize{\textit{1.5B}} & 53.71 & 56.10 & 47.65 & 47.21 \\
& PseudoReasoner (RoBERTa-large) & 78.33 & 78.91 & 79.69 & 80.27 \\
& PseudoReasoner (DeBERTa-v3-large) & 79.03 & 79.21 & 79.89 & 80.07 \\
& CAT (RoBERTa-large) \scriptsize{\textit{340M}} & 78.51 & 78.53 & 82.27 & 83.02 \\
& CAT (DeBERTa-v3-large) \scriptsize{\textit{435M}} & \underline{79.55} & \underline{79.39} & \underline{82.88} & \underline{83.52} \\
\midrule
\multirow{13}{*}{\begin{tabular}[c]{@{}l@{}}Large \\ Language \\ Models\end{tabular}} & ChatGPT (\texttt{openai/gpt-3.5-turbo}) & 69.29 & 68.65 & 68.54 & 68.12 \\ 
& \quad + Five-shot Exemplars & 69.42 & 70.40 & 70.27 & 72.08 \\
& \quad + Chain-of-thought & 74.82 & 72.32 & 71.48 & 72.85 \\
& LLAMA2 \scriptsize{\textit{7B}} & 46.29 & 43.90 & 40.81 & 41.25 \\
& \quad + Five-shot Exemplars & 47.92 & 44.89 & 74.67 & 76.80 \\
& LLAMA2 \scriptsize{\textit{13B}} & 48.17 & 48.59 & 48.31 & 48.55 \\
& \quad + Five-shot Exemplars & 49.29 & 49.90 & \underline{80.67} & 82.08 \\
& Mistral-v0.1 \scriptsize{\textit{7B}} & 46.29 & 43.90 & 58.09 & 58.07 \\
& \quad + Five-shot Exemplars & 51.00 & 50.06 & 65.09 & 69.80 \\
& LLAMA2 (LoRA Fine-tuned) \scriptsize{\textit{7B}} & \underline{75.80} & 76.27 & 79.89 & \underline{82.15} \\
& Mistral-v0.1 (LoRA Fine-tuned) \scriptsize{\textit{7B}} & 75.71 & \underline{76.76} & 79.59 & 80.35 \\
& VERA-T5 \scriptsize{\textit{5B}} & 70.76 & 70.29 & 72.60 & 76.85 \\
& VERA-T5 (Fine-tuned) \scriptsize{\textit{5B}} & 75.69 & 76.21 & 80.13 & 81.25 \\
\midrule
\multirow{5}{*}{\begin{tabular}[c]{@{}l@{}}\textbf{CANDLE} \\ \textbf{Distilled} \\ \textbf{\textit{(Ours)}}\end{tabular}} & RoBERTa-large \scriptsize{\textit{340M}} & 80.69$_{\uparrow2.18}$ & 80.99$_{\uparrow2.46}$ & 83.11$_{\uparrow0.84}$ & 84.50$_{\uparrow1.48}$ \\
& DeBERTa-v3-large \scriptsize{\textit{435M}} & \textbf{\underline{80.97}}$_{\uparrow1.42}$ & \textbf{\underline{81.14}}$_{\uparrow1.75}$ & \textbf{\underline{83.64}}$_{\uparrow0.76}$ & \textbf{\underline{84.64}}$_{\uparrow1.12}$ \\
& LLAMA2 (LoRA Fine-tuned) \scriptsize{\textit{7B}} & 77.48$_{\uparrow1.68}$ & 78.27$_{\uparrow2.00}$ & 81.68$_{\uparrow1.79}$ & 83.40$_{\uparrow1.25}$ \\
& Mistral-v0.1 (LoRA Fine-tuned) \scriptsize{\textit{7B}} & 77.77$_{\uparrow2.06}$ & 78.29$_{\uparrow1.53}$ & 81.95$_{\uparrow2.36}$ & 82.54$_{\uparrow2.19}$ \\
& VERA-T5 (Fine-tuned) \scriptsize{\textit{5B}} & 77.54$_{\uparrow1.85}$ & 78.03$_{\uparrow1.82}$ & 82.79$_{\uparrow2.66}$ & 83.61$_{\uparrow2.36}$ \\
\bottomrule
\end{tabular}
\vspace{-0.1in}
\caption{Performances (Accuracy\%) on CSKB conceptualization tasks. 
The best performances within each model type are \underline{underlined}, and the best among all models are \textbf{bold-faced}. 
}
\vspace{-0.15in}
\label{tab:CSKB_conceptualization_performance}
\end{table*}

\subsection{Downstream Applications}
\label{sec:downstream_applications}
In this section, we explore the downstream applications of CANDLE. 
By applying CANDLE to ATOMIC, the distilled conceptualizations and instantiations form a large-scale expansion of the original CSKB, which contains high-quality abstract and concrete commonsense knowledge. 
Leveraging both types of knowledge as supplementary training data, we enhance various downstream commonsense reasoning models.
Specifically, we utilize distilled conceptualizations in the CSKB conceptualization task~\cite{CAT}, while instantiations are used in generative commonsense inference (COMET;~\citealp{DBLP:conf/acl/BosselutRSMCC19}) and zero-shot commonsense QA tasks~\cite{DBLP:conf/aaai/MaIFBNO21}.
Due to space constraints, please refer to Appendix~\ref{appendix:task_setups},~\ref{appendix:dataset_description},~\ref{appendix:implementation_details} for task setups, dataset descriptions, and implementation details, respectively.

\subsubsection{CSKB Conceptualization}
\label{sec:CSKB_conceptualization}
\paragraph{Task Setup.}
The CSKB conceptualization task evaluates a model's ability to conceptualize a CSKB through two binary classification subtasks, which are crucial for performing CSKB conceptualization inference upon concept taxonomies~\cite{AbstractATOMIC}. 
The first subtask, event conceptualization, aims to determine whether $h_o$ can be correctly conceptualized using $h_a$, where $h_a$ is derived by replacing an instance $i \subset h_o$ with its linked concept $c$. 
The second subtask, triple conceptualization, aims to assess the plausibility of a conceptualized triple $(h_a, r, t)$ that represents abstract commonsense knowledge. 
Accuracy is used as the evaluation metric.
Following~\citet{CAT}, we use the AbstractATOMIC dataset provided by~\citet{AbstractATOMIC} as the evaluation benchmark.

To obtain our distilled models for these tasks, we first synthesize negative samples from CANDLE distilled conceptualizations. 
For event conceptualizations, a random concept from another head event without common words is selected as the negative candidate, while for triple conceptualization, a tail of another head event without common words under the same relation is selected. 
We then fine-tune language models on a balanced mixture of CANDLE distillation and synthesized negative samples to train two models, each serving as a pre-trained general discriminator in their respective task domain. 
These two models are subsequently fine-tuned on the training sets of AbstractATOMIC to fit into the benchmark, and their performances on the validation and test sets are reported.

\begin{table*}[t]
\small
\centering
\begin{tabular}{@{}l|cccccccc|c@{}}
\toprule
Training Data & Bleu-1 & Bleu-2 & Bleu-3 & Bleu-4 & METEOR & ROUGE-L & CIDEr & BERTScore & Human\\
\midrule
\multicolumn{8}{@{}l}{\textbf{Backbone: GPT2-XL~\cite{radford2019language} \scriptsize{\textit{1.5B}}}} \\
Zero-shot & 4.350 & 1.598 & 0.732 & 0.293 & 5.702 & 5.030 & 0.792 & 37.11 & 14.50 \\
ATOMIC & 45.72 & 29.18 & 21.12 & 16.15 & 29.97 & 49.69 & 64.61 & 76.09 & 70.50 \\
ATOMIC$^{20}_{20}$ & 42.15 & 25.77 & 17.82 & 13.14 & 29.82 & 47.61 & 63.70 & 70.39 & 76.50 \\
ATOMIC-10X & 45.38 & 29.20 & 21.09 & 16.15 & 30.09 & 49.86 & 65.02 & 75.89 & 77.50 \\
AbstractATOMIC & 45.30 & 29.08 & 21.00 & 16.06 & 29.98 & 48.61 & 63.98 & 75.56 & 71.50 \\
\textbf{CANDLE Distilled} & \textbf{\underline{50.71}} & \textbf{\underline{33.85}} & \textbf{\underline{25.55}} & \textbf{\underline{20.43}} & \textbf{\underline{32.45}} & \textbf{\underline{51.91}} & \textbf{\underline{69.68}} & \textbf{\underline{76.86}} & \underline{78.50} \\
\midrule
\multicolumn{8}{@{}l}{\textbf{Backbone: ChatGPT~\cite{openai2022chatgpt} (\texttt{openai/gpt-3.5-turbo})}} \\
Zero-shot & 11.82 & 4.258 & 1.891 & 0.926 & 13.87 & 13.73 & 4.350 & 49.28 & 78.50 \\
Five-shot & \underline{26.32} & \underline{12.50} & \underline{7.160} & \underline{4.415} & \underline{18.60} & \underline{24.65} & \underline{8.313} & \underline{58.69} & \textbf{\underline{81.00}} \\
Chain-of-thought & 9.906 & 3.568 & 1.556 & 0.736 & 11.85 & 11.02 & 2.905 & 46.17 & 64.00\\
\midrule
\multicolumn{8}{@{}l}{\textbf{Backbone: LLAMA2~\cite{LLAMA2} \scriptsize{\textit{7B}}}} \\
Zero-shot & 18.26 & 7.453 & 3.594 & 1.945 & 15.90 & 20.28 & 8.872 & 48.23 & 48.50 \\
Five-shot & 31.22 & 16.87 & 9.767 & 5.989 & 19.74 & 27.67 & 17.83 & 58.41 & 65.50 \\
ATOMIC & 42.04 & 23.01 & 14.10 & 9.125 & 27.80 & 42.90 & 53.17 & 71.52 & 68.50 \\
ATOMIC$^{20}_{20}$ & 41.07 & 22.46 & 13.62 & 8.619 & 27.74 & 42.42 & 53.28 & 71.77 & 74.00 \\
ATOMIC-10X & 42.07 & 23.08 & 14.14 & 9.198 & 28.14 & 42.75 & 53.69 & 71.93 & 76.50 \\
AbstractATOMIC & 42.78 & 23.64 & 14.58 & 9.471 & 27.74 & 42.55 & 53.12 & 71.51 & 71.00 \\
\textbf{CANDLE Distilled} & \underline{43.86} & \underline{24.40} & \underline{15.12} & \underline{10.00} & \underline{28.36} & \underline{43.86} & \underline{54.25} & \underline{72.94} & \underline{79.50} \\
\bottomrule
\end{tabular}
\vspace{-0.1in}
\caption{Performances (\%) of the commonsense inference modeling task (COMET) on the full test set of ATOMIC$^{20}_{20}$.
The best ones within each backbone are \underline{underlined}, and the best among all is \textbf{bold-faced}.
}
\vspace{-0.25in}
\label{tab:COMET_performance}
\end{table*}

\vspace{-0.1in}
\paragraph{Baselines.}
We evaluate our distilled models by comparing them against several baselines. 
These include supervised fine-tuned language models like RoBERTa-Large~\cite{DBLP:journals/corr/abs-1907-11692}, DeBERTa-V3-Large~\cite{he2023debertav}, GPT-2~\cite{radford2019language}, LLAMA2~\cite{LLAMA2}, Mistral~\cite{Mistral}, and VERA~\cite{VERA}, as well as semi-supervised methods such as PsuedoReasoner~\cite{DBLP:conf/emnlp/FangDZSWS22} and CAT~\cite{CAT}. 
Due to computational power limitations, we utilize LoRA~\cite{DBLP:conf/iclr/HuSWALWWC22} for fine-tuning LLMs. 
As additional baselines, we also consider prompting LLMs, including LLAMA2, Mistral, and ChatGPT. 
We explore both direct zero-shot prompting and alternative methods, such as with five-shot exemplars~\cite{DBLP:journals/corr/abs-2303-03846} and chain-of-thought reasoning~\cite{DBLP:conf/nips/Wei0SBIXCLZ22}.

\vspace{-0.1in}
\paragraph{Results and Analysis.}
Table~\ref{tab:CSKB_conceptualization_performance} shows the results.
CAT trained with DeBERTa-v3-large outperforms all other baselines for both tasks. 
Among LLMs, LLAMA and Mistral perform well after fine-tuning, but they struggle in prompting scenarios. 
However, pre-training on CANDLE's distilled conceptualizations consistently improves results for both tasks. 
For example, Mistral shows a significant improvement of 1.54\% and 2.19\% on two tasks compared to directly fine-tuning on AbstractATOMIC. 
Additionally, the distilled DeBERTa-v3-large surpasses all baseline models and achieves state-of-the-art performance. 
This can be attributed to the distilled conceptualizations obtained from CANDLE, which grant the model a more comprehensive understanding of conceptualizations and subsequently enhance its discriminatory capabilities.

\begin{table*}[t]
\small
\centering
\begin{tabular}{@{}l|c|lllll|l@{}}
\toprule
Model/Method & CSKB & a-NLI & CSQA & PIQA & SIQA & WG & Avg. \\ 
\midrule
\multicolumn{8}{@{}l}{\textbf{Pre-trained Language Models}} \\
RoBERTa-L~\cite{DBLP:journals/corr/abs-1907-11692} & - & 65.5 & 45.0 & 67.6 & 47.3 & 57.5 & 56.6 \\
DeBERTa-v3-L~\cite{he2023debertav} & - & 59.9 & 25.4 & 44.8 & 47.8 & 50.3 & 45.6 \\
Self-talk~\cite{DBLP:conf/emnlp/ShwartzWBBC20} & - & - & 32.4 & 70.2 & 46.2 & 54.7 & - \\
SMLM~\cite{DBLP:conf/emnlp/BanerjeeB20} & * & 65.3 & 38.8 & - & 48.5 & - & - \\
COMET-DynGen~\cite{DBLP:conf/aaai/BosselutBC21} & ATOMIC & - & - & - & 50.1 & - & - \\
MICO~\cite{DBLP:conf/emnlp/SuWFZSZ22} & ATOMIC & - & 44.2 & - & 56.0 & - & - \\
STL-Adapter~\cite{DBLP:conf/naacl/KimKKAHY22} & ATOMIC & 71.3 & 66.5 & 71.1 & 64.4 & 60.3 & 66.7 \\
DeBERTa-v3-L (MR)~\cite{DBLP:conf/aaai/MaIFBNO21} & ATM10X & 75.1 & 71.6 & 79.0 & 59.7 & 71.7 & 71.4 \\
DeBERTa-v3-L (MR)~\cite{DBLP:conf/aaai/MaIFBNO21} & ATOMIC & 76.0 & 67.0 & 78.0 & 62.1 & 76.0 & 71.8 \\
CAR-DeBERTa-v3-L~\cite{CAR} & ATOMIC & 78.9 & 67.2 & 78.6 & 63.8 & 78.1 & 73.3 \\
CAR-DeBERTa-v3-L~\cite{CAR} & AbsATM & \underline{79.6} & 69.3 & 78.6 & 64.0 & \underline{78.2} & \underline{73.9} \\
\textbf{DeBERTa-v3-L (CANDLE Distilled)} & CANDLE & \textbf{81.2}$_{\uparrow1.6}$ & 69.9$_{\uparrow0.6}$ & 80.3$_{\uparrow1.7}$ & 65.9$_{\uparrow1.9}$ & \textbf{78.3}$_{\uparrow0.1}$ & \textbf{74.9}$_{\uparrow1.0}$ \\
\midrule
\multicolumn{8}{@{}l}{\textbf{Large Language Models}} \\
GPT-3.5 (\texttt{text-davinci-003}) & - & 61.8 & 68.9 & 67.8 & 68.0 & 60.7 & 65.4 \\
ChatGPT (\texttt{gpt-3.5-turbo}) & - & 69.3 & 74.5 & 75.1 & 69.5 & 62.8 & 70.2 \\
\quad + Chain-of-thought & - & 70.5 & \underline{75.5} & 79.2 & \textbf{70.7} & 63.6 & 71.9 \\
\quad + Self-consistent chain-of-thought & - & 73.2 & \textbf{75.7} & \underline{81.7} & \underline{69.7} & 64.1 & 72.9 \\
GPT-4 (\texttt{gpt-4}) & - & 75.0 & 43.0 & 73.0 & 57.0 & 77.0 & 65.0 \\
LLAMA2 (7B;~\citealp{LLAMA2}) & - & 57.5 & 57.8 & 78.8 & 48.3 & 69.2 & 62.3 \\
LLAMA2 (13B;~\citealp{LLAMA2}) & - & 55.9 & 67.3 & 80.2 & 50.3 & 72.8 & 65.3 \\
Mistral-v0.1 (7B;~\citealp{Mistral}) & - & 51.0 & 59.6 & \textbf{83.0} & 42.9 & 75.3 & 62.4 \\
VERA-T5-xxl~\cite{VERA} & ATOMIC & 71.2 & 61.7 & 76.4 & 57.7 & 67.5 & 66.9 \\
VERA-T5-xxl~\cite{VERA} & ATM10X & 70.3 & 59.5 & 75.1 & 58.2 & 67.2 & 66.1\\
VERA-T5-xxl~\cite{VERA} & AbsATM & 73.2 & 63.0 & 77.2 & 58.1 & 68.1 & 68.0 \\
\textbf{VERA-T5-xxl (CANDLE Distilled)} & CANDLE & 73.8$_{\uparrow0.6}$ & 64.7$_{\uparrow1.7}$ & 77.6$_{\uparrow0.4}$ & 59.4$_{\uparrow1.2}$ & 71.3$_{\uparrow3.2}$ & 69.4$_{\uparrow1.4}$ \\
\bottomrule
\end{tabular}
\vspace{-0.1in}
\caption{Zero-shot evaluation results (Accuracy\%) on five commonsense question answering benchmarks. 
The best results are \textbf{bold-faced}, and the second-best ones are \underline{underlined}.
ATM10X stands for ATOMIC-10X~\cite{DBLP:conf/naacl/WestBHHJBLWC22} and AbsATM stands for AbstractATOMIC~\cite{AbstractATOMIC}.
}
\vspace{-0.15in}
\label{tab:csqa_performance}
\end{table*}

\subsubsection{Generative Commonsense Inference}
\label{sec:COMET_generation}
\paragraph{Task Setup.}
The task of generative commonsense inference modeling (COMET;~\citealp{DBLP:conf/acl/BosselutRSMCC19}) asks the model to generate commonsense tails $t$ based on given head $h_o$ and relation $r$ inputs. 
Following~\citet{DBLP:conf/aaai/HwangBBDSBC21}, we use the full test set of ATOMIC$^{20}_{20}$ as our evaluation benchmark.
We use several automatic metrics for evaluation, including BLEU~\cite{DBLP:conf/acl/PapineniRWZ02}, ROUGE-L~\cite{lin-2004-rouge}, METEOR~\cite{DBLP:conf/wmt/LavieA07}, CIDEr~\cite{DBLP:conf/cvpr/VedantamZP15}, and BERTScore~\cite{DBLP:conf/iclr/ZhangKWWA20}.
Meanwhile, four expert annotators are recruited to conduct expert evaluations of the generations. 
They are asked to annotate the plausibility of 200 randomly selected commonsense triple generations under each setting, and the resulting plausibility rates are reported.

Similar to training distilled models in previous tasks, we first pre-train GPT2 and LLAMA2-7B on critic-filtered CANDLE instantiations, where each $(h_{i'}, r, t)$ triple is concatenated into a sentence via natural language templates. 
Subsequently, we fine-tune these models on the training split of ATOMIC$^{20}_{20}$ to fit them into the benchmark. 
Finally, we report their performances on the test set.

\vspace{-0.1in}
\paragraph{Baselines.}
For baselines, we separately train GPT2 and LLAMA2-7B on the training sets of ATOMIC, ATOMIC$^{20}_{20}$, ATOMIC10X~\cite{DBLP:conf/naacl/WestBHHJBLWC22}, and AbstractATOMIC.
These models are then fine-tuned on the training split of ATOMIC$^{20}_{20}$ and evaluated on its test set.
We also include their zero-shot prompting performances, with LLAMA2 being evaluated with five-shot exemplars.
ChatGPT's performances under zero-shot, five-shot, and chain-of-thought settings are also reported.

\vspace{-0.1in}
\paragraph{Results and Analysis.}
Table~\ref{tab:COMET_performance} shows the results. 
Among the baselines, models pre-trained on ATOMIC-10X achieve the highest expert acceptance rate, surpassing those trained on AbstractATOMIC. 
This may be because ATOMIC-10X covers a wider range of commonsense relations consistent with ATOMIC$^{20}_{20}$. 
However, CANDLE distilled models achieve the highest scores compared to baselines with the same backbone model. 
For example, the CANDLE distilled LLAMA-7B model improves BERTScore by 1.01\% and expert-plausibility by 3.00\% compared to the best baseline. 
It also outperforms ChatGPT in all automatic metrics while maintaining a high plausibility rate of around 80\%.
This emphasizes the advantages of using CANDLE distilled instantiations for COMET training over traditional symbolic knowledge distillation methods or conceptualization augmentation.
Interestingly, we also observe that LLAMA2 has a tendency to generate long and contextually rich commonsense knowledge. 
On the other hand, GPT2, when fine-tuned on ATOMIC-like data, may generate shorter and more concise knowledge, which aligns with the format and length of knowledge in ATOMIC2020, thus achieving better results in automatic evaluations.
However, human annotators tend to consider long and contextually rich commonsense statements, generated by LLAMA2, as more plausible.

\subsubsection{Zero-shot Commonsense QA}
\label{sec:zero-shot commonsense QA}
\paragraph{Task Setup.}
The task of zero-shot commonsense QA involves selecting the most plausible option for commonsense questions without any supervision signals from benchmark data.
We follow the most effective pipeline by~\citet{DBLP:conf/aaai/MaIFBNO21}, which fine-tune language models on QA pairs synthesized from knowledge in CSKBs.
The head $h_o$ and relation $r$ of a $(h_o, r, t)$ triple are transformed into a question using natural language prompts, with the tail $t$ serving as the correct answer option. 
Distractors or negative examples are generated by randomly sampling tails from triples that do not share common keywords with the head. 
In addition to directly synthesizing from knowledge triples in ATOMIC, we augment ATOMIC by sampling triples from ATOMIC-10X, AbstractATOMIC, and CANDLE instantiations. 
The number of sampled triples is the same as in the original ATOMIC dataset.
We then synthesize them into QA pairs to train different baseline models and CANDLE distilled models. 
For our distilled models, we utilize QA pairs sourced from CANDLE-instantiation augmented ATOMIC to train a DeBERTa-v3-large model using the marginal ranking loss and a T5-xxl model~\cite{DBLP:journals/jmlr/RaffelSRLNMZLL20} following the training regime of VERA.
We evaluate the performance of all models on the validation split of Abductive NLI (aNLI;~\citealp{DBLP:conf/iclr/BhagavatulaBMSH20}), CommonsenseQA (CSQA;~\citealp{DBLP:conf/naacl/TalmorHLB19}), PhysicalIQA (PIQA;~\citealp{DBLP:conf/aaai/BiskZLGC20}), SocialIQA (SIQA;~\citealp{DBLP:conf/emnlp/SapRCBC19}), and WinoGrande (WG;~\citealp{DBLP:journals/cacm/SakaguchiBBC21}).
Accuracy is used as the evaluation metric.

\vspace{-0.1in}
\paragraph{Baselines.}
First, we report performances of vanilla RoBERTa-Large, DeBERTa-v3-Large, Self-talk~\cite{DBLP:conf/emnlp/ShwartzWBBC20}, COMET-DynaGen~\cite{DBLP:conf/aaai/BosselutBC21}, SMLM~\cite{DBLP:conf/emnlp/BanerjeeB20}, MICO~\cite{DBLP:conf/emnlp/SuWFZSZ22}, MR~\cite{DBLP:conf/aaai/MaIFBNO21}, STL-Adapter~\cite{DBLP:conf/naacl/KimKKAHY22}, and the previous state-of-the-art method, CAR~\cite{CAR}.
For MR and CAR, DeBERTa-v3-Large is used as the backbone, and their performances on ATOMIC-10X and AbstractATOMIC are also reported.
For LLMs, we report the performances of prompting GPT3.5~\cite{DBLP:conf/nips/BrownMRSKDNSSAA20}, ChatGPT, GPT4~\cite{GPT4}, LLAMA2, and Mistral in a zero-shot manner.
For ChatGPT, its performances with chain-of-thought~\cite{DBLP:conf/nips/Wei0SBIXCLZ22} and self-consistency chain-of-thought~\cite{DBLP:conf/iclr/0002WSLCNCZ23} prompting are also reported.
We also train several VERA-T5-xxl baselines on different sets of QA pairs as LLM baselines.

\vspace{-0.1in}
\paragraph{Results and Analysis.}
Table~\ref{tab:csqa_performance} shows the results, demonstrating that CANDLE distilled models generalize better than the baselines across several commonsense QA benchmarks.
For instance, VERA demonstrates an average improvement of 1.4\% compared to the best baseline.
This can be attributed to the inclusion of new entities and events in CANDLE instantiations that are absent in other CSKBs, where CANDLE instantiations can aid in answering commonsense questions that require knowledge of these new instances.
Furthermore, the distilled DeBERTa-v3-large model outperforms all baselines, including methods utilizing LLMs. 
This also indicates that augmenting with CANDLE distilled instantiations provides a more significant advantage compared to using symbolically distilled or abstract knowledge as training data.

\begin{table}[t]
\setlength{\tabcolsep}{4pt}
\small
\centering
\resizebox{!}{!}{
\begin{tabular}{@{}lll@{}}
\toprule
Critic & Conceptualization & Instantiation \\ \midrule
0.0 & 92.3\% & 85.5\% \\
0.5 & 94.6\% & 91.2\% \\
0.7 & 95.9\% & 93.3\% \\ 
0.9 & 98.3\% & 96.7\% \\
\bottomrule
\end{tabular}
}
\vspace{-0.1in}
\caption{Annotation results of distillations obtained from the second round of executing CANDLE.}
\vspace{-0.1in}
\label{tab:candle_second_roundstatistics}
\end{table}

\subsection{Analysis}
\subsubsection{Feasibility of Iterating CANDLE}
We first demonstrate the feasibility of iterating the CANDLE framework with more than one round. 
To do so, we randomly sample 10,000 distilled instantiations from LLAMA2 as the input for the CANDLE framework and execute the framework again, resulting in 200,000 conceptualizations and 200,000 instantiations. 
Subsequently, we randomly select 300 from each set and annotate them accordingly. 
The results are shown in Table~\ref{tab:candle_second_roundstatistics}. 
We observe that iterating the framework produces slightly better results than the first loop. 
This improvement may be attributed to the fact that the knowledge generated in the initial loop is more easily understood by LLMs compared to the human-annotated data in ATOMIC. 
Moreover, 58\% of the conceptualizations and 44\% of the instantiations are novel compared to the first loop. 
Based on these findings, we believe that our iterative framework is effective, and the iteration process enhances the augmentation of a CSKB through multiple iterations.

\begin{table}[t]
\setlength{\tabcolsep}{4pt}
\small
\centering
\resizebox{!}{!}{
\begin{tabular}{@{}llll@{}}
\toprule
Task & CSKB Concept. & COMET & CSQA \\
\midrule
Overlap Ratio & 10.1\% & 8.7\% & 5.3\% \\
Avg. Similarity & 0.39 & 0.38 & 0.31 \\
\bottomrule
\end{tabular}
}
\vspace{-0.1in}
\caption{Knowledge overlap ratio and average similarity between distilled knowledge and evaluation data.}
\vspace{-0.1in}
\label{tab:empirical_gain}
\end{table}

\subsubsection{Source of Empirical Gains}
Since LLAMA2 has been pre-trained on some evaluation benchmarks, it remains questionable whether the empirical gains in downstream tasks are due to knowledge overlap between distillations from LLAMA2 and the evaluation benchmarks.
To this extent, we further demonstrate that CANDLE distilled models perform better due to improved generalizability rather than relying on data overlap with the evaluation data.
We use SentenceBERT~\cite{DBLP:conf/emnlp/ReimersG19} to measure the textual similarity between the distilled knowledge and the evaluation data for each task. 
We then calculate the ratio of data that exhibits semantic overlap with a similarity score exceeding 0.5 and also report the average similarity. 
The results are shown in Table~\ref{tab:empirical_gain}.
Based on the results, we observe that the distilled knowledge has minimal overlap with the evaluation set. 
This indicates that the empirical gain primarily stems from our distilled knowledge, which improves the generalizability of the models, rather than relying on knowledge overlap with the evaluation sets.

\section{Conclusion}
This paper introduces CANDLE, a distillation framework that realizes the chain of conceptualization and instantiation over CSKBs.
We demonstrate the efficacy of CANDLE through comprehensive evaluations of the distilled knowledge and its positive impact on downstream tasks. 
Our research sheds light on distilling LLMs to enable more robust and generalizable commonsense reasoning.

\section*{Limitations}
The major limitation of CANDLE lies in the significant cost of distilling LLMs to obtain substantial knowledge. 
While the instantiation step of CANDLE utilizes the open foundation model LLAMA2, the conceptualization is still performed by ChatGPT due to the unsatisfactory performance of other open-source LLMs and the high quality of ChatGPT's generation. 
Consequently, a considerable amount of funding is required to distill conceptualizations for CANDLE to function effectively. 
Future works should address this issue by exploring advanced prompting methods or employing stronger open-source LLMs as the foundation for distilling conceptualizations.

Furthermore, it should be noted that CANDLE has only been validated on ATOMIC. 
However, CANDLE is not limited to any specific format of commonsense knowledge, allowing it to operate on any CSKB. 
Future research can address this by extending the evaluation of CANDLE to other CSKBs and conducting follow-up experiments to explore their benefits on more downstream tasks.

Another interesting direction to investigate is utilizing the chain of conceptualization and instantiation as a foundation for enhancing weak-to-strong generalization~\cite{DBLP:journals/corr/abs-2312-09390}.
By conceptualizing and instantiating weak supervision data, we can generate more robust and generalized training signals, which ultimately strengthens the learning process.
This can also be effectively incorporated into the training process of self-rewarding language models~\cite{yuan2024selfrewarding}.

\section*{Ethics Statement}
To avoid generating harmful or unethical content from LLMs like ChatGPT and LLAMA2, we recruit four expert annotators, who are graduate and undergraduate students specializing in machine commonsense in natural language processing, to verify the ethics and potential harm of the generated content. 
A thorough assessment of a random sample has been conducted, and no significant harm has been identified.
All training and evaluation datasets used are publicly available and shared under open-access licenses solely for research purposes, aligning with their intended usage. 
These datasets have been carefully anonymized and desensitized to protect data privacy and confidentiality.
The expert annotators involved in this study are fully aware of the annotation protocol and the intended use of their annotations. 
Their participation in this research is voluntary, and they have agreed to contribute without receiving any compensation.
Thus, the authors believe that this paper does not raise any ethical concerns.

\section*{Acknowledgements}
We thank the anonymous reviewers and the area chair for their constructive comments.
The authors of this paper were supported by the NSFC Fund (U20B2053) from the NSFC of China, the RIF (R6020-19 and R6021-20), and the GRF (16211520 and 16205322) from RGC of Hong Kong. 
We also thank the support from the UGC Research Matching Grants (RMGS20EG01-D, RMGS20CR11, RMGS20CR12, RMGS20EG19, RMGS20EG21, RMGS23CR05, RMGS23EG08). 

\bibliography{custom}
\bibliographystyle{acl_natbib}

\appendix

\begin{center}
    {\Large\textbf{Appendices}}
\end{center}

\section{Distillation Details}
\label{appendix:distillation_prompts}
This section provides additional details about the CANDLE distillation process not covered in the main body text. 
First, we present the prompts used to instruct ChatGPT to perform contextualized conceptualizations and LLAMA2 to perform contextualized instantiation. 
For prompting ChatGPT to distill conceptualizations, we use a few-shot prompt as shown below:

\begin{displayquote}
\textbf{\texttt{\small Following the given examples, you are required to conceptualize the instance (enclosed by []) in the last given event into abstract concepts. The concept should still fit into the instance's original sentence. Make sure that the generated abstract concepts are general and not simply hypernyms of the instance.}}\\
\ldots \\
\textbf{\textcolor{headcolor}{\texttt{\small Event <i>: PersonX enjoys drinking in the [bar], as a result, PersonX feels relaxed. [bar] can be conceptualized as }}\textcolor{tailcolor}{\texttt{\small\textbf{Social Gathering Place}}}}\\
\ldots \\
\textbf{\textcolor{headcolor}{\texttt{\small Event <N>: PersonX likes [painting on the beach], as a result, PersonX will go to the beach. [painting on the beach] can be conceptualized as}}}
\end{displayquote}

Similarly, for prompting LLAMA2-13B to distill instantiations based on previously generated conceptualizations, we use a few-shot prompt as shown below:

\begin{displayquote}
\textbf{\texttt{\small Following the given examples, you are required to instantiate the concept (enclosed by []) in the last given event into entities or events. If the event only contains the concept, then instantiate it to an event starting with a subject PersonX or PersonY. If the event contains other words, then instantiate it to an entity. The instance should still fit into the original sentence. Make sure that the generated instance is specific.}}\\
\ldots \\
\textbf{\textcolor{headcolor}{\texttt{\small Event <i>: PersonX enjoys drinking in the [Social Gathering Place], as a result, PersonX feels relaxed. [Social Gathering Place] can be instantiated as }}\textcolor{tailcolor}{\texttt{\small\textbf{beer festival}}}}\\
\ldots \\
\textbf{\textcolor{headcolor}{\texttt{\small Event <N>: PersonX likes [exercise], as a result, PersonX will go to the stadium. [exercise] can be conceptualized as}}}
\end{displayquote}

\begin{table}[t]
\centering
\small
\begin{tabular}{@{}l|cccc@{}}
\toprule
 & Abs.ATM & CANDLE \\ 
\midrule
\#Unq. event & 15,388 & 15,359 \\
\#Unq. instance & 21,493 & 21,442 \\
\#Unq. conceptualization & 31,227 & \textbf{853,499} \\
\#Tot. conceptualization & 503,588 & \textbf{6,181,391} \\
\#Unq. instantiation & - & \textbf{676,737} \\
\#Tot. instantiation & - & \textbf{6,181,391} \\
\midrule
Avg. \#concept/event & 32.73 & \textbf{173.33} \\
Avg. \#Unq. concept/event & 28.33 & \textbf{167.76} \\
Avg. \#concept/instance & 23.43 & \textbf{124.16} \\ 
Avg. \#Unq. concept/instance & 17.27 & \textbf{100.88} \\
\bottomrule
\end{tabular}
\caption{Statistics of conceptualizations and instantiations in AbstractATOMIC (Abs.ATM;~\citealp{AbstractATOMIC}) and CANDLE.
Tot. stands for total, Unq. stands for unique, and Avg. stands for average.}
\label{tab:additional_AbsATMconcept_statistics}
\end{table}

\begin{table}[t]
\centering
\small
\begin{tabular}{@{}l|ccc@{}}
\toprule
Relation & ATOMIC & Abs.ATM & CANDLE \\ 
\midrule
xEffect & 78,832 & 938,330 & 964,765 \\
oEffect & 28,351 & 333,845 & 346,363 \\
xWant & 101,249 & 1,170,835 & 1,322,810 \\
oWant & 43,079 & 484,570 & 551,391 \\
xReact & 62,969 & 510,476 & 480,259 \\
oReact & 26,570 & 224,706 & 208,538 \\
xNeed & 74,272 & 900,429 & 894,338 \\
xAttr & 110,791 & 838,191 & 810,958 \\
xIntent & 45,490 & 519,813 & 601,969 \\
\midrule
Total & 572,053 & 5,921,195 & 6,181,391 \\
\bottomrule
\end{tabular}
\caption{Statistics of abstract commonsense knowledge triples by relations in ATOMIC, AbstractATOMIC (Abs.ATM;~\citealp{AbstractATOMIC}), and CANDLE.}
\label{tab:additional_AbsATMtriple_statistics}
\end{table}

These prompts are consistent with our descriptions in Section~\ref{sec:contextualized_conceptualization} and Section~\ref{sec:contextualized_instantiation}, where the task description is first presented, followed by human-authored examples, and finally, the event we want to conceptualize or instantiate.
We also leverage several tricks in the prompt, such as numbering the examples, generating concepts instead of hypernyms, and keeping the generated responses concise.
Finally, we parse the generations via manually defined rules and compile them into a dataset.

Additionally, we introduce some generation settings when prompting LLMs. 
For ChatGPT, we access it through the official OpenAI APIs\footnote{\href{https://chat.openai.com/}{https://chat.openai.com/}}. 
The code of the accessed version is \texttt{gpt-3.5-turbo-0613}. 
We set the temperature to 1.0 and the maximum length for generated tokens to 200. 
To conceptualize all events in ATOMIC into 20 conceptualizations each, the time required for the distillation process is approximately ten days and the financial budget is around 1500 USD.

For LLAMA2, we access it via the Huggingface Library~\cite{DBLP:conf/emnlp/WolfDSCDMCRLFDS20}. 
The code of the accessed model is \texttt{meta-llama/Llama-2-13b-chat-hf}\footnote{\href{https://huggingface.co/meta-llama/Llama-2-13b-chat-hf}{https://huggingface.co/meta-llama/Llama-2-13b-chat-hf}}. 
When prompting, we use the Top-k sampling decoding strategy and set $k = 10$. 
We set the maximum length of generated tokens to 200. 
The models are hosted on sixteen NVIDIA-V100 GPUs, and the time required to distill the entire dataset is approximately one month.

After collecting 20 conceptualizations for every head event in ATOMIC and further instantiating them to new entities and events, we construct an expanded knowledge base of ATOMIC. 
We also include more statistics, as shown in Table~\ref{tab:additional_AbsATMconcept_statistics} and Table~\ref{tab:additional_AbsATMtriple_statistics}. 
For instantiations, they share the same relational distribution as abstract commonsense triples since we only instantiate them once.
These statistics indicate that, compared to AbstractATOMIC, which is the only available conceptualization benchmark based on ATOMIC, CANDLE contains more abstract commonsense triples and many more unique conceptualizations. 
According to our results, it can also be expected that the abstract knowledge distilled from CANDLE is of better quality than AbstractATOMIC, which human annotations or any filtering have not verified.

For critic filtering, we use the state-of-the-art conceptualization discriminator developed by~\citet{CAT}. 
This discriminator is utilized to assess the plausibility of CANDLE distilled conceptualizations. 
It considers the original event, the instance being conceptualized, and the target concept as its inputs and generates a score ranging from 0 to 1 to represent plausibility.
For instantiation, we use the pre-trained VERA model released by~\citet{VERA}. 
We convert the instantiated commonsense knowledge triple into a declarative statement and request an estimation of its plausibility from VERA. 
This estimation is provided as a score ranging from 0 to 1.
The output scores from both models serve as the critical values assigned to each CANDLE distillation. 
These critical values are then subjected to further filtering based on various thresholds.

Additionally, following~\citet{AbsPyramid}, we calculate the percentage of unique abstract concepts using BLEU soft uniqueness~\cite{DBLP:conf/sigir/ZhuLZGZWY18,DBLP:conf/naacl/WestBHHJBLWC22}. 
We define a concept, denoted as $x$, as unique if $BLEU_1(C, x) < 0.5$, where $C$ represents all concepts that share the same head event and identified instance with $x$ in AbstractATOMIC.
Here, 0.5 serves as an empirical threshold. 
Our distillation process yields 92.3\% unique conceptualizations, indicating a significantly higher diversity than previous datasets.

Similarly, we evaluate the uniqueness of the newly introduced head events resulting from our chain of conceptualization and instantiation. 
To determine uniqueness, we define an instantiated head event, referred to as $h_{i'}$, as unique if $BLEU_1(h_o, h_{i'}) < 0.5$, where $h_o$ represents the original head event in ATOMIC. 
The threshold of 0.5 is an empirical threshold. 
Our empirical results demonstrate that 78.6\% of the instantiated events are unique compared to ATOMIC, highlighting the effectiveness of CANDLE in enhancing the semantic coverage of the CSKB.


\section{Task Setups}
\label{appendix:task_setups}
\subsection{CSKB Conceptualization}
We follow the task definition of~\citet{AbstractATOMIC} and~\citet{CAT} to formulate the CSKB conceptualization task.
Specifically, conceptualizing an event-centric CSKB to derive abstract commonsense knowledge comprises two steps~\cite{AbstractATOMIC}: event conceptualization and triple conceptualization, which correspond to two subtasks studied in this paper.
Denote the triples in the original CSKB as $D_o=\{(h_o,r,t)|h_o \in H_o, r \in R, t \in T\}$, where $H_o$, $R$, and $T$ are the set of heads, relations, and tails in the original CSKB. 
The first step only operates on head events without considering the context in $r$ and $t$. 
The goal of event conceptualization is to produce a conceptualized head event $h_a$ from the original head $h_o$ to represent an abstraction of $h_o$.
In the second step, the task is to verify whether the conceptualized head $h_a$ still makes sense in the context of $r$ and $t$, as $r$ and $t$ will further restrict the level of abstractness in $h_a$.
Plausible $(h_a,r,t)$ triples will be considered as valid abstract commonsense knowledge.
By enhancing the performance of discriminative models on these tasks, they can function as more precise critic filters and automate the conceptualization process of a CSKB when linked to concept taxonomies.

\subsection{Generative Commonsense Inference}
The task of generative commonsense inference was studied by both~\citet{DBLP:conf/acl/BosselutRSMCC19} and~\citet{DBLP:conf/aaai/HwangBBDSBC21}.
It requires a generative model to complete the tail $t$ of a commonsense assertion based on a given pair of head $h$ and commonsense relation $r$.
In this paper, we follow~\citet{DBLP:conf/aaai/HwangBBDSBC21} and use ATOMIC$^{20}_{20}$ as the evaluation benchmark, in which the full testing set is used for model evaluation.
The task of COMET is important in the domain of commonsense as it serves as a fundamental component for numerous high-level applications that necessitate commonsense reasoning, such as zero-shot commonsense question answering with self-talk~\cite{DBLP:conf/emnlp/ShwartzWBBC20} and dynamic graph construction~\cite{DBLP:conf/aaai/BosselutBC21}, narrative reasoning~\cite{peng-etal-2022-inferring}, and dialogue generation~\cite{tu-etal-2022-misc}. 
Improving COMET can potentially benefit other domains that require commonsense understanding.

\subsection{Zero-shot Commonsense QA}
The task of zero-shot commonsense QA evaluates a model's reasoning generalizability on unseen QA entries without any supervision signals from the corresponding annotated training data.
Several methods have been proposed to tackle this task, including those by~\citet{DBLP:conf/emnlp/ShwartzWBBC20,DBLP:conf/aaai/BosselutBC21,DBLP:conf/naacl/KimKKAHY22,DBLP:conf/emnlp/ShiWFX0LS23}. 
The most effective pipeline, as suggested by~\citet{DBLP:conf/aaai/MaIFBNO21}, injects commonsense knowledge into language models via fine-tuning on QA pairs synthesized from knowledge in CSKBs.
During the fine-tuning process, the head $h_o$ and relation $r$ of a $(h_o, r, t)$ triple from a CSKB are transformed into a question using natural language prompts, with the tail $t$ serving as the correct answer option. 
Distractors or negative examples are generated by randomly sampling tails from triples that do not share common keywords with the head. 
This fine-tuning procedure enhances the model's knowledge not only for QA benchmarks constructed from CSKBs but also improves its ability to answer unseen commonsense questions in a more generalized manner.
In this paper, we follow the task definition, model training, and model evaluation pipeline by~\citet{DBLP:conf/aaai/MaIFBNO21} to study the impact of distilling student models from CANDLE instantiations.
For baselines, we compare models trained on QA pairs synthesized from ATOMIC, ATOMIC-10X, and AbstractATOMIC.
For ATOMIC-10X, 0.9 is used as the critic filtering threshold.

\begin{table}[t]
\centering
\small
\begin{tabular}{@{}l|l|ccc@{}}
\toprule
Data & Type & Train & Dev & Test \\ 
\midrule
\multirow{2}{*}{$D^l$} & \#event & 107,384 & 12,117 & 11,503 \\
 & \#triple & 65,386 & 8,403 & 7,408 \\ 
 \midrule
\multirow{2}{*}{$D^u$} & \#event & 304,983 & 36,023 & 31,578 \\
 & \#triple & 4,851,272 & 499,523 & 570,400 \\ 
 \bottomrule
\end{tabular}
\caption{Statistics of labeled data $D^l$ and unlabeled data $D^u$ in AbstractATOMIC.}
\label{tab:appendix_AbstractAtomic_statistics}
\end{table}

\section{Dataset Descriptions}
\label{appendix:dataset_description}
This section covers additional details and statistics of datasets and benchmarks used in downstream task evaluations.

\subsection{CSKB Conceptualization}
In CSKB Conceptualization tasks, we use the AbstractATOMIC~\cite{AbstractATOMIC} dataset as the evaluation benchmark.
It is a benchmark dataset built upon ATOMIC and consists of event conceptualization data and abstract knowledge triples. 
The event conceptualizations are based on head events in ATOMIC, identified through syntactic parsing and matching with rules to search for concept candidates in Probase~\cite{DBLP:conf/sigmod/WuLWZ12} and WordNet~\cite{DBLP:journals/cacm/Miller95}. 
The abstract knowledge triples connect conceptualized head events with their non-abstract counterparts from ATOMIC, forming commonsense knowledge at the concept level.
Human annotations are used to verify the correctness of some conceptualizations and their resulting abstract commonsense triples. 
In total, 131K conceptualizations of 7K (45\%) ATOMIC head events and 81K (1.3\%) conceptualized triples are manually annotated, with a large number remaining unlabeled.
The data is partitioned by following ATOMIC's original split of head events.
Detailed statistics are shown in Table~\ref{tab:appendix_AbstractAtomic_statistics}.
In this paper, we evaluate all models using the test set from the annotated subset as the evaluation data. 
Meanwhile, we obtain CANDLE distilled models using the training set from the annotated subset to fine-tune discriminative models pre-trained on CANDLE conceptualizations.
Supervised baselines are trained on the training set of AbstractATOMIC, while semi-supervised baselines also leverage the unlabeled data.

\subsection{Generative Commonsense Inference}
To evaluate COMET, we adopt the same evaluation setting employed by~\citet{DBLP:conf/aaai/HwangBBDSBC21} for assessing commonsense generative models on the ATOMIC$^{20}_{20}$ dataset's test set. 
We use the entire test set, consisting of 34,689 triples across 23 different commonsense relations, to ensure the robustness of the evaluation. 
Additionally, we use the full training set to fine-tune models that were pre-trained on various CSKBs and CANDLE instantiations to fit them into the benchmark.

Recently,~\citet{DBLP:conf/emnlp/West0SLJLCHBB023} successfully trained a powerful commonsense inference generator using an open-format symbolic knowledge distillation framework. 
Once they release their data and models, we will incorporate them as another baseline in our comparisons.

\begin{table}[t]
\centering
\small
\begin{tabular}{@{}l|ccccc@{}}
\toprule
 & aNLI & CSQA & PIQA & SIQA & WG \\ 
\midrule
\#QA Pairs & 1,532 & 1,221 & 1,838 & 1,954 & 1,267 \\
\#Options & 2 & 5 & 2 & 3 & 2\\
\bottomrule
\end{tabular}
\caption{Statistics on the number of QA pairs and the number of options for each question in benchmarks used in the zero-shot commonsense QA task.}
\label{tab:appendix_qa_benchmark_statistics}
\end{table}

\subsection{Zero-shot Commonsense QA}
We follow~\citet{DBLP:conf/aaai/MaIFBNO21,CAR,DBLP:conf/emnlp/ShiWFX0LS23} and use the validation split of five commonsense QA benchmarks: Abductive NLI (aNLI;~\citealp{DBLP:conf/iclr/BhagavatulaBMSH20}), CommonsenseQA (CSQA;~\citealp{DBLP:conf/naacl/TalmorHLB19}), PhysicalIQA (PIQA;~\citealp{DBLP:conf/aaai/BiskZLGC20}), SocialIQA (SIQA;~\citealp{DBLP:conf/emnlp/SapRCBC19}), and WinoGrande (WG;~\citealp{DBLP:journals/cacm/SakaguchiBBC21}).
These benchmarks evaluate different aspects, including abductive reasoning, concept-level commonsense reasoning, physical commonsense understanding, emotional and social commonsense reasoning, and pronoun resolution.
The validation splits are used as the official test sets may not be publicly available.
Statistics on the number of QA pairs and the number of options per question are reported in Table~\ref{tab:appendix_qa_benchmark_statistics}.

\begin{table*}[t]
\centering
\small
\begin{tabular}{@{}l|l@{}}
\toprule
Task & Prompt \\ 
\midrule
Event. & \begin{tabular}[c]{@{}l@{}}Given the event ``PersonX enjoys drinking in the bar,'' can ``bar'' be conceptualized as ``entertainment venue''?\\ Here, conceptualized means represented by a general concept. Answer 'Yes' or 'No' only without any other word.\end{tabular} \\ \midrule
Triple. & \begin{tabular}[c]{@{}l@{}}Given the assertion: PersonX enjoys drinking in entertainment venue, as a result, PersonX feels relaxed.\\ entertainment venue is a general concept and represents many possible instances.\\ Is this assertion plausible? Answer 'Yes' or 'No' only without any other word.\end{tabular} \\ \midrule
COMET & \begin{tabular}[c]{@{}l@{}}Please complete the given commonsense assertion with a few words. Don't extend writing afterward.\\ PersonX hears strange noises, as a result, PersonX will\end{tabular} \\ \midrule
aNLI & \begin{tabular}[c]{@{}l@{}}Premise: Jim decided to be a rockstar.\\ Choice A: but didn't know how to play an instrument. Jim signed up for guitar lessons.\\ Choice B: Jim knew he would need to have a nickname. Jim signed up for guitar lessons.\\ Which one is more likely to happen, given the premise? Only answer A or B without any other word.\end{tabular} \\ \midrule
CSQA & \begin{tabular}[c]{@{}l@{}}Question: He was at the gym trying to build muscle, what is it called that he is trying to build muscle on?\\ Choice A: body of animal \\ Choice B: arm \\ Choice C: bodybuilder \\ Choice D: body of dog \\ Choice E: human body \\ Which choice is correct? Only answer A or B or C or D or E without any other word.\end{tabular} \\ \midrule
PIQA & \begin{tabular}[c]{@{}l@{}}Goal: To remove an avocado from the shell\\ Choice A: cut the avocado lengthwise, remove the pit, and scoop with a spoon\\ Choice B: cut the avocado width wise, remove the pit, and scoop with a spoon\\ Which choice can achieve the goal? Only answer A or B without any other word.\end{tabular} \\ \midrule
SIQA & \begin{tabular}[c]{@{}l@{}}Question: Robin went to the polls and posted her ballot for the candidate she wanted. \\ As a result, Robin wanted to:\\ Choice A: bomb the candidate \\ Choice B: attend a rally \\ Choice C: go home.\\ Which choice is correct? Only answer A or B or C without any other word.\end{tabular} \\ \midrule
WG & \begin{tabular}[c]{@{}l@{}}Question: Jessica enjoyed a simple, basic life with Betty, but\\ Choice A: Jessica was bored having a quiet existence.\\ Choice B: Betty was bored having a quiet existence.\\ Which choice is correct? Only answer A or B without any other word.\end{tabular} \\ \bottomrule
\end{tabular}
\caption{Prompts used for evaluating LLM baselines across various tasks in a zero-shot scenario. Event. stands for event conceptualization discrimination and Triple. stands for triple conceptualization discrimination.}
\label{tab:appendix_baselines_prompt}
\end{table*}

\section{Implementation Details}
\label{appendix:implementation_details}
This section provides additional implementation details in downstream task evaluations.

First, we use the Huggingface\footnote{\href{https://huggingface.co/}{https://huggingface.co/}} Library~\cite{DBLP:conf/emnlp/WolfDSCDMCRLFDS20} to build all models.
We reproduce all baselines according to implementation details described in their original papers.
The reported results are consistent with their original papers if the same experiment is included.
For CANDLE distilled models, please refer to the subsections (Appendix~\ref{appendix:implementation_cskb_conceptualization}, \ref{appendix:implementation_comet}, \ref{appendix:implementation_commonsense_qa}) below.

For methods involving LLMs, we use their instruction fine-tuned versions as the backbone for the baselines. 
For LLAMA2, the accessed version is \texttt{meta-llama/Llama-2-7b/13b-chat-hf}.
For Mistral, we use \texttt{mistralai/Mistral} \texttt{-7B-Instruct-v0.1}.
This remains consistent whether we prompt them directly or fine-tune them for downstream tasks, as we have observed that the instruction-finetuned versions generally result in better performance.
For ChatGPT, we access it through Microsoft Azure APIs\footnote{\href{https://azure.microsoft.com/en-us/products/ai-services/}{https://azure.microsoft.com/en-us/products/ai-services/}}.
The code of the accessed version for ChatGPT is \texttt{gpt-35-turbo-20230515}, and for GPT4 is \texttt{gpt-4-20230515}.
The maximum generation length is set to 100 tokens for all tasks.
For fine-tuning LLAMA2 and Mistral, we use the open code base of LLaMa-Factory\footnote{\href{https://github.com/hiyouga/LLaMA-Factory}{https://github.com/hiyouga/LLaMA-Factory}}.
Please refer to the subsections below for hyperparameter settings.

All experiments are conducted on sixteen NVIDIA-V100 (32G) GPUs.

For baselines involving prompting LLMs, we follow the approach done by~\citet{DBLP:conf/iclr/RobinsonW23} and~\citet{DBLP:journals/corr/abs-2304-14827}, where each task is formulated in either a generative format or as multiple-choice QA.
Table~\ref{tab:appendix_baselines_prompt} shows the prompts used in zero-shot prompting scenarios.
To incorporate five-shot exemplars, we include five randomly selected examples from the training set of each benchmark. 
These examples are merged into the prompt using the same format as the question, with the addition of including the answer at the end.
For chain-of-thought reasoning, we prompt LLM in a two-step inference process. 
In the first step, we delve deeper into the question by requesting an intermediate-step rationale. 
Then, in the second step, we seek an answer based on the question and the previous step's response by asking LLM to answer ``Yes or No only'' or select the correct option from a set of answers directly.

\begin{table*}[t]
\small
\centering
\begin{tabular}{@{}l|cccccccc@{}}
\toprule
Training Data & Bleu-1 & Bleu-2 & Bleu-3 & Bleu-4 & METEOR & ROUGE-L & CIDEr & BERTScore \\
\midrule
\multicolumn{8}{@{}l}{\textbf{Backbone: GPT2-XL~\cite{radford2019language} \scriptsize{\textit{1.5B}}}} \\
Zero-shot & 4.350 & 1.598 & 0.732 & 0.293 & 5.702 & 5.030 & 0.792 & 37.11 \\
ATOMIC & 32.23 & 19.06 & 13.27 & 10.28 & 17.63 & 25.50 & 20.15 & 58.39   \\
\quad + Finetune & 45.72 & 29.18 & 21.12 & 16.15 & 29.97 & 49.69 & 64.61 & 76.09   \\
ATOMIC$^{20}_{20}$ & 42.15 & 25.77 & 17.82 & 13.14 & 29.82 & 47.61 & 63.70 & 70.39  \\
ATOMIC-10X & 33.69 & 18.82 & 11.71 & 7.910 & 18.78 & 25.69 & 19.29 & 61.47   \\
\quad + Finetune & 45.38 & 29.20 & 21.09 & 16.15 & 30.09 & 49.86 & 65.02 & 75.89   \\
AbstractATOMIC & 29.46 & 17.16 & 11.89 & 9.019 & 17.42 & 24.30 & 19.95 & 57.83   \\
\quad + Finetune & 45.30 & 29.08 & 21.00 & 16.06 & 29.98 & 48.61 & 63.98 & 75.56   \\
\textbf{CANDLE Distilled} & 26.91 & 16.44 & 12.31 & 10.28 & 17.66 & 23.66 & 21.36 & 57.15   \\
\quad + Finetune & \textbf{\underline{50.71}} & \textbf{\underline{33.85}} & \textbf{\underline{25.55}} & \textbf{\underline{20.43}} & \textbf{\underline{32.45}} & \textbf{\underline{51.91}} & \textbf{\underline{69.68}} & \textbf{\underline{76.86}}  \\
\midrule
\multicolumn{8}{@{}l}{\textbf{Backbone: ChatGPT~\cite{openai2022chatgpt} (\texttt{openai/gpt-3.5-turbo})}} \\
Zero-shot & 11.82 & 4.258 & 1.891 & 0.926 & 13.87 & 13.73 & 4.350 & 49.28   \\
Five-shot & \underline{26.32} & \underline{12.50} & \underline{7.160} & \underline{4.415} & \underline{18.60} & \underline{24.65} & \underline{8.313} & \underline{58.69}   \\
Chain-of-thought & 9.906 & 3.568 & 1.556 & 0.736 & 11.85 & 11.02 & 2.905 & 46.17  \\
\midrule
\multicolumn{8}{@{}l}{\textbf{Backbone: LLAMA2~\cite{LLAMA2} \scriptsize{\textit{7B}}}} \\
Zero-shot & 18.26 & 7.453 & 3.594 & 1.945 & 15.90 & 20.28 & 8.872 & 48.23   \\
Five-shot & 31.22 & 16.87 & 9.767 & 5.989 & 19.74 & 27.67 & 17.83 & 58.41   \\
ATOMIC & 29.94 & 16.44 & 10.03 & 6.631 & 19.02 & 25.75 & 18.71 & 59.68   \\
\quad + Finetune & 42.04 & 23.01 & 14.10 & 9.125 & 27.80 & 42.90 & 53.17 & 71.52   \\
ATOMIC$^{20}_{20}$ & 41.07 & 22.46 & 13.62 & 8.619 & 27.74 & 42.42 & 53.28 & 71.77   \\
ATOMIC-10X & 33.06 & 17.65 & 9.986 & 6.078 & 19.22 & 25.32 & 17.80 & 61.25  \\
\quad + Finetune & 42.07 & 23.08 & 14.14 & 9.198 & 28.14 & 42.75 & 53.69 & 71.93   \\
AbstractATOMIC & 26.08 & 13.27 & 7.799 & 5.018 & 15.08 & 21.20 & 14.78 & 56.83   \\
\quad + Finetune & 42.78 & 23.64 & 14.58 & 9.471 & 27.74 & 42.55 & 53.12 & 71.51   \\
\textbf{CANDLE Distilled} & 28.93 & 15.56 & 9.468 & 6.140 & 18.60 & 25.37 & 17.20 & 60.27   \\
\quad + Finetune & \underline{43.86} & \underline{24.40} & \underline{15.12} & \underline{10.00} & \underline{28.36} & \underline{43.86} & \underline{54.25} & \underline{72.94}   \\
\bottomrule
\end{tabular}
\caption{Full performances (\%) of the commonsense inference modeling task (COMET) on the full test set of ATOMIC$^{20}_{20}$~\cite{DBLP:conf/aaai/HwangBBDSBC21}.
The best performances using each backbone are \underline{underlined}, and the best among all backbones are \textbf{bold-faced}.
Finetune refers to fine-tuning back on the training set of ATOMIC$^{20}_{20}$.
}
\vspace{-0.1in}
\label{tab:COMET_performance_full}
\end{table*}

\subsection{CSKB Conceptualization}
\label{appendix:implementation_cskb_conceptualization}
For RoBERTa and DeBERTa-v3, we use a learning rate of 5e-6 and a batch size of 64. 
To optimize the models, we use an AdamW optimizer~\cite{DBLP:conf/iclr/LoshchilovH19} and evaluate the model's performance every 25 steps. 
The maximum sequence lengths for the tokenizers are set to 25 and 35 for the two discriminative subtasks, respectively.
Early stopping is used where the best checkpoint is selected when the largest validation accuracy is achieved.
The models are trained on CANDLE distillation for one epoch and fine-tuned on the training set of AbstractATOMIC for one epoch.

For LLMs, such as LLAMA2 and Mistral, we use LoRA for fine-tuning, and the LoRA rank and $\alpha$ are set to 64 and 64.
We use an Adam~\cite{DBLP:journals/corr/KingmaB14} optimizer with a learning rate of 5e-6 and a batch size of 64.
The models are fine-tuned for two epochs, and the checkpoint with the highest validation set accuracy is selected.
All experiments are repeated three times using different random seeds, and the average performances are reported.

For VERA, we follow the exact same implementation\footnote{\href{https://github.com/liujch1998/vera}{https://github.com/liujch1998/vera}} as released by~\citet{VERA}.
To transform our binary classification subtasks into declarative formats, we begin by converting each piece of data into a declarative sentence using predefined natural language templates. 
Next, we create a corresponding negative statement by simply incorporating the word ``not'' into the correct sentence. 
For instance, a pair of statements is: ``PersonX enjoys drinking at the bar. Bar is a social gathering place.'' and ``PersonX enjoys drinking at the bar. Bar is not a social gathering place.''
The accessed backbone model is \texttt{liujch1998/vera}, and all other hyperparameter settings follow the default implementation.
The model is trained on CANDLE distillation for one epoch and then fine-tuned on AbstractATOMIC for another.

\subsection{Generative Commonsense Inference}
\label{appendix:implementation_comet}
For COMET, we implement the open-sourced code by~\citet{DBLP:conf/aaai/HwangBBDSBC21} as our base to fine-tune the GPT2 model.
The model is first pre-trained on CANDLE instantiations for one epoch, followed by fine-tuning on ATOMIC$^{20}_{20}$ for another epoch.
An Adam~\cite{DBLP:journals/corr/KingmaB14} optimizer is used with a learning rate of 1e-5 and a batch size of 32.
A linear scheduler is used to decrease the learning rate gradually.

For LLAMA2-7B, we fine-tune it with the DeepSpeed framework~\cite{DBLP:conf/sc/AminabadiRALLZRSZRH22} by using FP16 as the precision.
We optimize the model with an Adam~\cite{DBLP:journals/corr/KingmaB14} optimizer with a learning rate of 1e-4 and a batch size of 64.
The maximum length for the input and generated sentence concatenation is 500.
We warm up the model with 3000 steps and evaluate the model every 1000 steps.
A linear scheduler is also used.
The LoRA rank is set to 8, and the $\alpha$ is set to 32.

In Table~\ref{tab:COMET_performance_full}, we present supplementary automatic evaluation results, including models that have been pre-trained solely on CSKBs and CANDLE instantiations without subsequent fine-tuning.

\begin{table*}[t]
\small
\centering
\resizebox{\linewidth}{!}{
\begin{tabular}{l|l|c}
\toprule
Original & Concept./Instant. & Critic \\ 
\midrule
\multirow{4}{*}{\begin{tabular}[c]{@{}l@{}}PersonX swims in the lake,\\ \textit{as a result, PersonX feels},\\ tired.\end{tabular}} 
& PersonX swims in \textbf{\textcolor{concept_color}{freshwater}}, \textit{as a result, PersonX feels}, tired. & 0.97 \\ 
& PersonX swims in \textbf{\textcolor{instance_color}{the sea}}, \textit{as a result, PersonX feels}, tired. & 0.87 \\
\cmidrule(l){2-3}
 & PersonX \textbf{\textcolor{concept_color}{swims}}, \textit{as a result, PersonX feels}, tired. & 0.89 \\
 & PersonX \textbf{\textcolor{instance_color}{swims every week}}, \textit{as a result, PersonX feels}, tired. & 0.81 \\
\midrule
\multirow{4}{*}{\begin{tabular}[c]{@{}l@{}}PersonX is sitting in class,\\ \textit{as a result, PersonX will},\\ learns something.\end{tabular}} 
& PersonX is sitting in \textbf{\textcolor{concept_color}{instructional period}}, \textit{as a result, PersonX will}, learns something. & 0.54 \\ 
& PersonX is sitting in \textbf{\textcolor{instance_color}{a math class}}, \textit{as a result, PersonX will}, learns something. & 0.75 \\
\cmidrule(l){2-3}
 & PersonX \textbf{\textcolor{concept_color}{study}}, \textit{as a result, PersonX will}, learns something. & 0.78 \\
 & PersonX \textbf{\textcolor{instance_color}{learns how to do the exam}}, \textit{as a result, PersonX will}, learns something. & 0.81 \\
\midrule
\multirow{4}{*}{\begin{tabular}[c]{@{}l@{}}PersonX buys PersonY a gift,\\ \textit{as a result, PersonY feels},\\ joyful.\end{tabular}} 
& \textbf{\textcolor{concept_color}{remembrance}}, \textit{as a result, PersonY feels}, joyful. & 0.19 \\ 
& \textbf{\textcolor{instance_color}{PersonX reminisce}}, \textit{as a result, PersonY feels}, joyful. & 0.27 \\
\cmidrule(l){2-3}
 & PersonX \textbf{\textcolor{concept_color}{shopping}}, \textit{as a result, PersonY feels}, joyful. & 0.61 \\
 & PersonX \textbf{\textcolor{instance_color}{buys a new toy for PersonY}}, \textit{as a result, PersonY feels}, joyful. & 0.90 \\
\midrule
\multirow{4}{*}{\begin{tabular}[c]{@{}l@{}}PersonX always fought,\\ \textit{as a result, PersonY feels},\\ angry.\end{tabular}} 
& PersonX always \textbf{\textcolor{concept_color}{violent behavior}}, \textit{as a result, PersonY feels}, angry. & 0.98 \\ 
& PersonX always \textbf{\textcolor{instance_color}{punch others hardly}}, \textit{as a result, PersonY feels}, angry. & 0.91 \\
\cmidrule(l){2-3}
 & \textbf{\textcolor{concept_color}{combative personality}}, \textit{as a result, PersonY feels}, angry. & 0.98 \\
 & PersonX \textbf{\textcolor{instance_color}{PersonX likes to join a fight}}, \textit{as a result, PersonY feels}, angry. & 0.85 \\
\midrule
\multirow{4}{*}{\begin{tabular}[c]{@{}l@{}}PersonX gets a new bike,\\ \textit{as a result, PersonX wants},\\ to ride it.\end{tabular}} 
& PersonX gets a \textbf{\textcolor{concept_color}{transportation tool}}, \textit{as a result, PersonX wants}, to ride it. & 0.92 \\ 
& PersonX gets a \textbf{\textcolor{instance_color}{motor}}, \textit{as a result, PersonX wants}, to ride it. & 0.98 \\
\cmidrule(l){2-3}
 & \textbf{\textcolor{concept_color}{bike possession}}, \textit{as a result, PersonX wants}, to ride it. & 0.93 \\
 & \textbf{\textcolor{instance_color}{PersonX has a nice bicycle}}, \textit{as a result, PersonX wants}, to ride it. & 0.89 \\
\midrule
\multirow{4}{*}{\begin{tabular}[c]{@{}l@{}}PersonX spends time with PersonY,\\ \textit{PersonX is seen as},\\ social.\end{tabular}} 
& PersonX spends \textbf{\textcolor{concept_color}{love-building period}} with PersonY, \textit{PersonX is seen as}, social. & 0.05 \\ 
& PersonX spends \textbf{\textcolor{instance_color}{time in love}} with PersonY, \textit{PersonX is seen as}, social. & 0.37 \\
\cmidrule(l){2-3}
 & \textbf{\textcolor{concept_color}{social activity}}, \textit{PersonX is seen as}, social. & 0.64 \\
 & \textbf{\textcolor{instance_color}{PersonX enjoys going to parties}}, \textit{PersonX is seen as}, social. & 0.73 \\
\midrule
\multirow{4}{*}{\begin{tabular}[c]{@{}l@{}}PersonX hears sirens,\\ \textit{as a result, PersonX will},\\ make way to the siren.\end{tabular}}
& \textbf{\textcolor{concept_color}{emergency response}}, \textit{as a result, PersonX will}, make way to the siren. & 0.37 \\ 
& \textbf{\textcolor{instance_color}{PersonX sees an ambulance coming}}, \textit{as a result, PersonX will}, make way to the siren. & 0.74 \\
\cmidrule(l){2-3}
 & PersonX hears \textbf{\textcolor{concept_color}{loud noise}}, \textit{as a result, PersonX will}, make way to the siren. & 0.67 \\
 & PersonX hears \textbf{\textcolor{instance_color}{a fire truck beeping}}, \textit{as a result, PersonX will}, make way to the siren. & 0.77 \\
 \bottomrule
\end{tabular}
}
\caption{Case studies of \textbf{\textcolor{concept_color}{conceptualizations}} and \textbf{\textcolor{instance_color}{instantiations}} distilled from CANDLE in their original context.
Original stands for the original triple sampled from ATOMIC.
In the Concept./Instant. column, each box contains an abstract commonsense triple that includes \textbf{\textcolor{concept_color}{conceptualization}}, followed by an instantiated commonsense triple with \textbf{\textcolor{instance_color}{instantiation}}.
We demonstrate two ways to conceptualize each original triple from ATOMIC.
}
\label{tab:appendix_case_study}
\end{table*}

\begin{table*}[t]
\small
\centering
\begin{tabular}{@{}l|c|lllll|l@{}}
\toprule
Model/Method & CSKB & a-NLI & CSQA & PIQA & SIQA & WG & Avg. \\ 
\midrule
\multicolumn{8}{@{}l}{\textbf{Pre-trained Language Models}} \\
Random Vote & - & 50.0 & 20.0 & 50.0 & 33.3 & 50.0 & 40.7 \\
Majority Vote & - & 50.8 & 20.9 & 50.5 & 33.6 & 50.4 & 41.2 \\
GPT2-L~\cite{radford2019language} & - & 56.5 & 41.4 & 68.9 & 44.6 & 53.2 & 52.9 \\
RoBERTa-L~\cite{DBLP:journals/corr/abs-1907-11692} & - & 65.5 & 45.0 & 67.6 & 47.3 & 57.5 & 56.6 \\
DeBERTa-v3-L~\cite{he2023debertav} & - & 59.9 & 25.4 & 44.8 & 47.8 & 50.3 & 45.6 \\
Self-talk~\cite{DBLP:conf/emnlp/ShwartzWBBC20} & - & - & 32.4 & 70.2 & 46.2 & 54.7 & - \\
SMLM~\cite{DBLP:conf/emnlp/BanerjeeB20} & * & 65.3 & 38.8 & - & 48.5 & - & - \\
COMET-DynGen~\cite{DBLP:conf/aaai/BosselutBC21} & ATOMIC & - & - & - & 50.1 & - & - \\
MICO~\cite{DBLP:conf/emnlp/SuWFZSZ22} & ATOMIC & - & 44.2 & - & 56.0 & - & - \\
STL-PLM~\cite{DBLP:conf/naacl/KimKKAHY22} & ATOMIC & 71.6 & 64.0 & 72.2 & 63.2 & 60.5 & 66.3 \\
MTL~\cite{DBLP:conf/naacl/KimKKAHY22} & CWWV & 69.6 & 67.3 & 72.5 & 52.0 & 57.2 & 63.7 \\
MTL~\cite{DBLP:conf/naacl/KimKKAHY22} & CSKG & 69.8 & 67.1 & 72.0 & 61.9 & 59.3 & 66.0 \\
STL-Adapter~\cite{DBLP:conf/naacl/KimKKAHY22} & ATOMIC & 71.3 & 66.5 & 71.1 & 64.4 & 60.3 & 66.7 \\
STL-Adapter~\cite{DBLP:conf/naacl/KimKKAHY22} & CSKG & 71.5 & 66.7 & 72.1 & 64.7 & 59.0 & 66.8 \\
RoBERTa-L (MR)~\cite{DBLP:conf/aaai/MaIFBNO21} & ATM$_{10X}$ & 70.8 & 64.2 & 71.7 & 61.0 & 60.7 & 65.7 \\
RoBERTa-L (MR)~\cite{DBLP:conf/aaai/MaIFBNO21} & ATOMIC & 70.8 & 64.2 & 72.1 & 63.1 & 59.2 & 65.9 \\
RoBERTa-L (MR)~\cite{DBLP:conf/aaai/MaIFBNO21} & CWWV & 70.0 & 67.9 & 72.0 & 54.8 & 59.4 & 64.8 \\
RoBERTa-L (MR)~\cite{DBLP:conf/aaai/MaIFBNO21} & CSKG & 70.5 & 67.4 & 72.4 & 63.2 & 60.9 & 66.8 \\
DeBERTa-v3-L (MR)~\cite{DBLP:conf/aaai/MaIFBNO21} & ATM10X & 75.1 & 71.6 & 79.0 & 59.7 & 71.7 & 71.4 \\
DeBERTa-v3-L (MR)~\cite{DBLP:conf/aaai/MaIFBNO21} & ATOMIC & 76.0 & 67.0 & 78.0 & 62.1 & 76.0 & 71.8 \\
ZS-Fusion~\cite{DBLP:conf/naacl/KimKKAHY22} & CWWV & 69.6 & 67.6 & 73.1 & 53.7 & 59.5 & 64.7 \\
ZS-Fusion~\cite{DBLP:conf/naacl/KimKKAHY22} & CSKG & 72.4 & 68.3 & 73.0 & 66.7 & 60.9 & 68.3 \\
MKIF~\cite{DBLP:journals/corr/abs-2305-05936} & CSKG & 72.5 & 71.0 & 73.1 & - & 61.0 & - \\
CAR-RoBERTa-L~\cite{CAR} & ATOMIC & 72.3 & 64.8 & 73.2 & 64.8 & 61.3 & 67.3 \\
CAR-RoBERTa-L~\cite{CAR} & AbsATM & 72.7 & 66.3 & 73.2 & 64.0 & 62.0 & 67.6 \\
CAR-DeBERTa-v3-L~\cite{CAR} & ATOMIC & 78.9 & 67.2 & 78.6 & 63.8 & 78.1 & 73.3 \\
CAR-DeBERTa-v3-L~\cite{CAR} & AbsATM & \underline{79.6} & 69.3 & 78.6 & 64.0 & \underline{78.2} & \underline{73.9} \\
\textbf{DeBERTa-v3-L (CANDLE Distilled)} & CANDLE & \textbf{81.2}$_{\uparrow1.6}$ & 69.9$_{\uparrow0.6}$ & 80.3$_{\uparrow1.7}$ & 65.9$_{\uparrow1.9}$ & \textbf{78.3}$_{\uparrow0.1}$ & \textbf{74.9}$_{\uparrow1.0}$ \\
\midrule
\multicolumn{8}{@{}l}{\textbf{Large Language Models}} \\
GPT-3.5 (\texttt{text-davinci-003}) & - & 61.8 & 68.9 & 67.8 & 68.0 & 60.7 & 65.4 \\
ChatGPT (\texttt{gpt-3.5-turbo}) & - & 69.3 & 74.5 & 75.1 & 69.5 & 62.8 & 70.2 \\
\quad + Chain-of-thought & - & 70.5 & \underline{75.5} & 79.2 & \textbf{70.7} & 63.6 & 71.9 \\
\quad + Self-consistent chain-of-thought & - & 73.2 & \textbf{75.7} & \underline{81.7} & \underline{69.7} & 64.1 & 72.9 \\
GPT-4 (\texttt{gpt-4}) & - & 75.0 & 43.0 & 73.0 & 57.0 & 77.0 & 65.0 \\
LLAMA2 (7B;~\citealp{LLAMA2}) & - & 57.5 & 57.8 & 78.8 & 48.3 & 69.2 & 62.3 \\
LLAMA2 (13B;~\citealp{LLAMA2}) & - & 55.9 & 67.3 & 80.2 & 50.3 & 72.8 & 65.3 \\
Mistral-v0.1 (7B;~\citealp{Mistral}) & - & 51.0 & 59.6 & \textbf{83.0} & 42.9 & 75.3 & 62.4 \\
VERA-T5-xxl~\cite{VERA} & ATOMIC & 71.2 & 61.7 & 76.4 & 57.7 & 67.5 & 66.9 \\
VERA-T5-xxl~\cite{VERA} & ATM10X & 70.3 & 59.5 & 75.1 & 58.2 & 67.2 & 66.1\\
VERA-T5-xxl~\cite{VERA} & AbsATM & 73.2 & 63.0 & 77.2 & 58.1 & 68.1 & 68.0 \\
\textbf{VERA-T5-xxl (CANDLE Distilled)} & CANDLE & 73.8$_{\uparrow0.6}$ & 64.7$_{\uparrow1.7}$ & 77.6$_{\uparrow0.4}$ & 59.4$_{\uparrow1.2}$ & 71.3$_{\uparrow3.2}$ & 69.4$_{\uparrow1.4}$ \\
\midrule
\multicolumn{8}{@{}l}{\textbf{Supervised Learning \& Human Performance}} \\
RoBERTa-L (Supervised) & - & 85.6 & 78.5 & 79.2 & 76.6 & 79.3 & 79.8 \\
DeBERTa-v3-L (Supervised) & - & 89.0 & 82.1 & 84.5 & 80.1 & 84.1 & 84.0 \\
VERA-T5 (Multitask Supervised) & - & 83.9 & 77.8 & 88.5 & 80.1 & 92.4 & 84.5 \\
Human Performance & - & 91.4 & 88.9 & 94.9 & 86.9 & 94.1 & 91.2 \\
\bottomrule
\end{tabular}
\caption{Full zero-shot evaluation results (Accuracy\%) on five commonsense question answering benchmarks. 
The best results are \textbf{bold-faced}, and the second-best ones are \underline{underlined}.
$\uparrow$ signifies the improvement CANDLE-distilled models achieve compared to the best baseline with the same backbone model.
ATM10X stands for ATOMIC-10X~\cite{DBLP:conf/naacl/WestBHHJBLWC22} and AbsATM stands for AbstractATOMIC~\cite{AbstractATOMIC}.
All scores are retrieved from their original papers.
For the GPT-X series, some results are retrieved from~\citet{DBLP:conf/emnlp/West0SLJLCHBB023}.
}
\label{tab:csqa_performance_full}
\end{table*}

\subsection{Zero-shot Commonsense QA}
\label{appendix:implementation_commonsense_qa}
For the task of zero-shot commonsense QA, we adopt the code base provided by~\citet{CAR}\footnote{\href{https://github.com/HKUST-KnowComp/CAR}{https://github.com/HKUST-KnowComp/CAR}} and~\citet{VERA}\footnote{\href{https://github.com/liujch1998/vera}{https://github.com/liujch1998/vera}} to train two CANDLE distilled models.
All hyperparameters and optimization strategies are kept unchanged from their original implementations as default settings. 
The models are trained for two epochs using QA pairs obtained from augmented-ATOMIC, including augmentations from ATOMIC-10X, AbstractATOMIC, and CANDLE instantiations.

Meanwhile, we present a comprehensive table presenting the results of all current methodologies for the task of zero-shot commonsense QA in Table~\ref{tab:csqa_performance_full}.
Notably, our CANDLE distilled models continue to exhibit strong performance compared to other models pre-trained on QA pairs sourced from multiple CSKBs. 
This serves as compelling evidence for the efficacy of CANDLE.

\begin{figure}[t]
     \centering
     \includegraphics[width=1\linewidth]{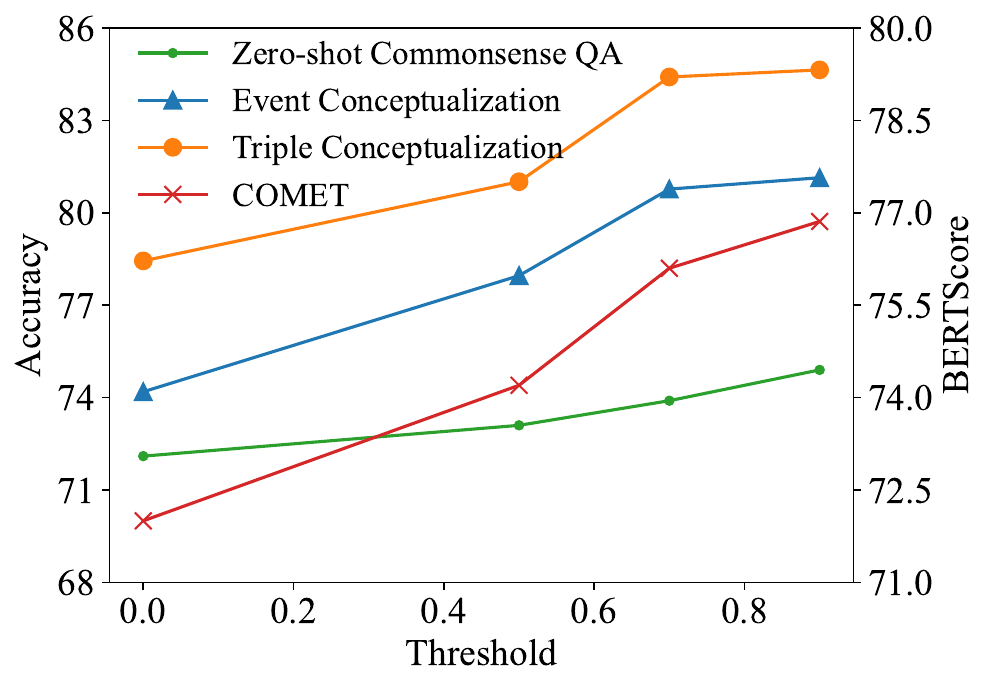}
     \vspace{-0.2in}
     \caption{Ablation results examining the impact of different threshold values in CANDLE's critic filtering.}
     \vspace{-0.2in}
    \label{fig:appendix_ablation}
\end{figure}

\section{Annotation Details}
\label{appendix:annotation}
This paper utilizes expert annotations to assess the quality of distilled conceptualizations and instantiations, as well as evaluate the generations of different models for the COMET downstream task.
We follow the annotation setting of~\citet{DBLP:conf/acl/YuWLBSLG0Y23,DBLP:journals/corr/abs-2402-18169,DBLP:conf/emnlp/ChengQCFWCRGZSZ23}, where four graduate students with ample experience in natural language processing research and expertise in commonsense reasoning are recruited as expert annotators to carry out the annotations. 
Their participation in the annotation process is voluntary and unpaid, in accordance with local laws, and is considered a contribution to this paper.
Detailed instructions are provided to the annotators for each task, ensuring that they understand the requirements thoroughly. 
For CANDLE distillation evaluation, the annotators are asked to determine (1) the correctness of the conceptualizations and instantiations and (2) the plausibility of their formed triples. 
For COMET generation evaluation, they are asked to determine the plausibility of the generated triples.
For each question, we also highlight the part to be considered by the annotators for their convenience.
The annotation process is conducted independently, without any internal discussions among the annotators regarding the results. 
For each task, two annotators independently vote for each triple, and only when both annotators provide a positive vote will the triple be considered accepted or plausible.
To prevent bias and ensure impartial results for CANDLE, the task input is randomly shuffled during the annotation process. 
As a result, the expert annotators achieve a pairwise agreement (IAA;~\citealp{landis1977application}) of 0.80 and a Fleiss-kappa~\cite{fleiss1971measuring} of 0.61, indicating a remarkably high level of internal agreement.

\section{Ablation Study}
\label{appendix:ablation_study}
In this section, we examine the impact of our critic filters on the ablation of CAN-DLE. 
Specifically, we investigate the effect of different levels of critic threshold or completely abandoning critic filtering on downstream tasks. 
We conduct four experiments with different settings, denoted as $t\in \{0, 0.5, 0.7, 0.9\}$, where $t$ = 0 corresponds to abandoning critic filtering and using all distilled knowledge as complementary training data.
For detailed statistics, please refer to Table~\ref{tab:candle_statistics}.
For each value of $t$, we select the distilled knowledge with a critic score higher than $t$ and utilize it as complementary training data to train student models for the three downstream tasks. 
We employ the same training strategies described in the main body of the paper. 
In the case of CSKB conceptualization and zero-shot commonsense QA tasks, we utilize DeBERTa-v3-large as the backbone model, with accuracy as the evaluation metric. 
For COMET, we use GPT2 and evaluate using the BERTScore as the evaluation metric. 
The results are visualized in Figure~\ref{fig:appendix_ablation}.
Our analysis reveals a consistent trend where higher threshold values yield improved performance, indicating the reliability of our critic filter. 
However, it is worth noting that setting the threshold above 0.9 may potentially lead to even better performance. 
Nevertheless, such a trade-off comes with a downside: it reduces the amount of usable knowledge in each distillation round, which can impede the iterative process. 
The reason for this is that when the number of distilled conceptualizations and instantiations decreases significantly in each round, CANDLE is unable to incorporate new instantiated data for future distillation iterations. 
As a result, the ``convergence'' of those high-critic data occurs prematurely in CANDLE.

\section{Case Study}
\label{appendix:case_study}
We present some examples in Table~\ref{tab:appendix_case_study} to show conceptualizations and instantiations generated by CANDLE, along with their corresponding critic values assigned by our critic-filtering discriminators.
It can be observed that both ChatGPT and LLAMA2 exhibit the ability to generate high-quality knowledge based on given instructions. 
Furthermore, they can introduce novel conceptualizations and events during the distillation chain, effectively meeting our expectations of CANDLE.
Future works can investigate the feasibility of incorporating conceptualization and abstract knowledge into more downstream tasks, such as complex reasoning~\cite{DBLP:journals/corr/abs-2403-07398,DBLP:conf/aaai/ZhangZMWS0SZ24,DBLP:conf/nips/BaiLW0S23} and commonsense knowledge graph denoising~\cite{DBLP:conf/emnlp/DengW0LS23}. 

\end{document}